\documentclass{ieeeaccess}
\usepackage{cite}
\usepackage{amsmath,amssymb,amsfonts}
\usepackage{algorithmic}
\usepackage{graphicx}
\usepackage{textcomp}

\usepackage{subcaption}
\usepackage{booktabs}
\usepackage{algorithm}
\usepackage{textgreek}
\usepackage{enumitem}
\usepackage{comment}
\makeatletter
\@namedef{ver@everyshi.sty}{}
\makeatother
\usepackage{natbib}
\setcitestyle{square,numbers}
\graphicspath{ {./img/} }

\usepackage[capitalize]{cleveref}
\crefname{section}{Sec.}{Secs.}
\Crefname{section}{Section}{Sections}
\Crefname{table}{Table}{Tables}
\crefname{table}{Tab.}{Tabs.}

\makeatletter
\renewcommand\tiny{\@setfontsize\tiny{7.8}{8.8}}
\makeatother

\setcitestyle{comma}

\begin{document}
\history{Date of publication xxxx 00, 0000, date of current version xxxx 00, 0000.}
\doi{10.1109/ACCESS.2017.DOI}

\title{AccelAT: A Framework for Accelerating the Adversarial Training of Deep Neural Networks through Accuracy Gradient}
\author{\uppercase{FARZAD NIKFAM}\authorrefmark{1}, \IEEEmembership{Graduate Student Member, IEEE},
\uppercase{ALBERTO MARCHISIO}\authorrefmark{2}, \IEEEmembership{Graduate Student Member, IEEE}, \uppercase{MAURIZIO MARTINA}\authorrefmark{1}, \IEEEmembership{Senior Member, IEEE}, 
\uppercase{and MUHAMMAD SHAFIQUE}\authorrefmark{3}, \IEEEmembership{Senior Member, IEEE}}
\address[1]{Department of Electrical, Electronics and Telecommunication Engineering, Politecnico di Torino, 10129 Torino TO, Italy; name.surname@polito.it}
\address[2]{Institute of Computer Engineering, Technische Universität Wien (TU Wien), 1040 Vienna, Austria; alberto.marchisio@tuwien.ac.at}
\address[3]{eBrain Lab, Division of Engineering, New York University Abu Dhabi, UAE; muhammad.shafique@nyu.edu}
\tfootnote{
This work has been supported in part by the Doctoral College Resilient Embedded Systems, which is run jointly by the TU Wien’s Faculty of Informatics and the UAS Technikum Wien. This work was also supported in parts by the NYUAD Center for Interacting Urban Networks (CITIES), funded by Tamkeen under the NYUAD Research Institute Award CG001, Center for Cyber Security (CCS), funded by Tamkeen under the NYUAD Research Institute Award G1104, and Center for Artificial Intelligence and Robotics (CAIR), funded by Tamkeen under the NYUAD Research Institute Award CG010.}

\markboth
{F. Nikfam \headeretal: AccelAT: A Framework for Accelerating the Adversarial Training of Deep Neural Networks through Accuracy Gradient}
{F. Nikfam \headeretal: AccelAT: A Framework for Accelerating the Adversarial Training of Deep Neural Networks through Accuracy Gradient}

\corresp{Corresponding author: Farzad Nikfam (e-mail: farzad.nikfam@polito.it).}

\begin{abstract}
Adversarial training is exploited to develop a robust Deep Neural Network (DNN) model against the malicious altered data. These attacks may have catastrophic effects on DNN models but are indistinguishable for a human being. For example, an external attack can modify an image adding noises invisible for a human eye, but a DNN model misclassifies
the image.
A key objective for developing robust DNN models is to use a learning algorithm that is fast but can also give model that is robust against different types of adversarial attacks. Especially for adversarial training, enormously long training times are needed for obtaining high accuracy under many different types of adversarial samples generated using different adversarial attack techniques.

This paper aims at accelerating the adversarial training to enable fast development of robust DNN models against adversarial attacks. The general method for improving the training performance is the hyperparameters fine-tuning, where the learning rate is one of the most crucial hyperparameters. By modifying its shape (the value over time) and value during the training, we can obtain a model robust to adversarial attacks faster than standard training. 

First, we conduct experiments on two different datasets (CIFAR10, CIFAR100), exploring various techniques. Then, this analysis is leveraged to develop a novel fast training methodology, \textit{AccelAT}, which automatically adjusts the learning rate for different epochs based on the accuracy gradient. The experiments show comparable results with the related works, and in several experiments, the adversarial training of DNNs using our \textit{AccelAT} framework is conducted up to 2$\times$ faster than the existing techniques. Thus, our findings boost the speed of adversarial training in an era in which security and performance are fundamental optimization objectives in DNN-based applications. To facilitate reproducible research this is the AccelAT open-source framework: https://github.com/Nikfam/AccelAT.
\end{abstract}


\titlepgskip=-15pt

\maketitle

\section{Introduction}
\label{sec:introduction}
Machine Learning (ML)~\cite{DBLP:journals/tccn/Simeone18, DBLP:journals/corr/abs-1708-08296, ml_raschka_2017} is an ever-expanding field and has achieved wide proliferation in recent years due to the development of highly-efficient hardware, such as GPUs.
Nevertheless, the advanced and complex ML models require gigantic training time. Therefore, its acceleration not only has a direct impact on the usage of large GPU-based datacenters in which typically the training is conducted but will also enable training on low-cost multi-GPU workstations, as well as will ease the development of new research directions, such as continuous learning. On the other hand, in the last decade, it has been discovered that models are highly vulnerable to external attacks, and nowadays, the ML models need to be robust against such attacks to be deployed in safety-critical applications.

\subsection{Target Research Problem and Challenges}
Adversarial training~\cite{DBLP:conf/iclr/MadryMSTV18} has become a popular method for training Deep Neural Networks (DNNs)~\cite{Goodfellow-et-al-2016} with robustness against the adversarial attacks. Unfortunately, robust DNNs are not always easy to be trained, as it takes 3$\times$ to 30$\times$ longer time~\cite{DBLP:conf/nips/ShafahiNG0DSDTG19} to obtain high accuracy when adversarial (noisy) samples are added to the training dataset, compared to the standard (i.e., non adversarial) training. Hence, it is essential to create DNN models that are not only robust but also quite fast to be trained. To obtain the above-discussed properties, we propose to employ fast training techniques for advanced adversarial training of DNNs, and show the feasibility of this design strategy by designing a novel fast training method, called \textit{AccelAT}.

\begin{figure}[b]
    \centering
    \includegraphics[width=1\linewidth]{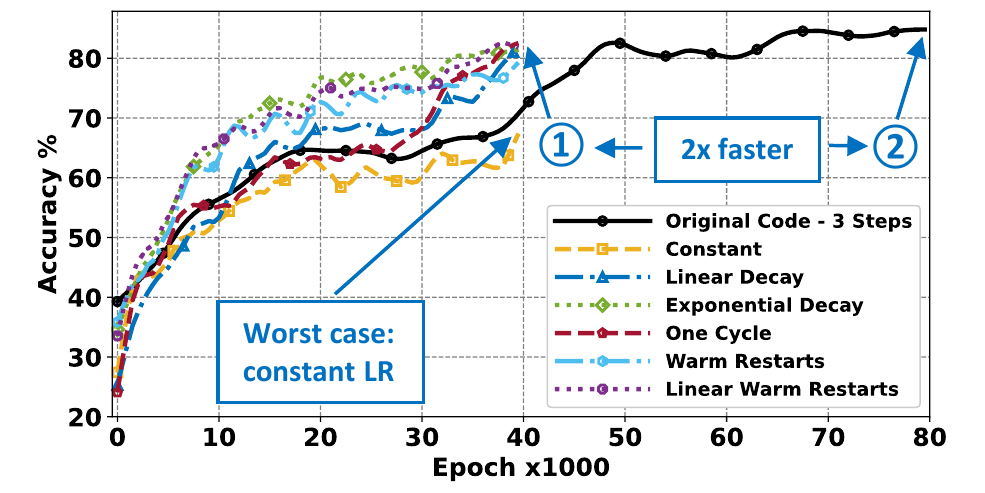}
    \caption{FAT Training of ResNet-20 on the \textit{CIFAR10} natural images dataset with different LR techniques.}
    \label{fig:figure_FAT_cifar10}
\end{figure}

\subsection{Motivational Case Study}
Nowadays, advanced methods for adversarial training - such as, Free Adversarial Training (FAT)~\cite{DBLP:conf/nips/ShafahiNG0DSDTG19}, YOPO~\cite{DBLP:conf/nips/ZhangZLZ019}, and Trades~\cite{DBLP:conf/icml/ZhangYJXGJ19} - are currently used to obtain DNN models which are robust against adversarial attacks. In our analyses, we focus on the FAT method, which is already highly optimized compared to the original adversarial training method~\cite{DBLP:conf/iclr/MadryMSTV18}. 
To further accelerate the training process, we employ various fast training techniques, focusing mainly on the study of hyperparameters. Since the learning rate (LR) has a strong influence on the convergence of the DNN training process, we analyze and how its variation affects the training speed for accurate and robust DNN models. The fast training techniques analyzed are \textit{linear decay}, \textit{exponential decay}, \textit{one cycle}~\cite{DBLP:journals/corr/abs-1803-09820}, and \textit{warm restarts}~\cite{DBLP:conf/iclr/LoshchilovH17,DBLP:conf/tencon/MishraS19}. The results applied to the ResNet-20 DNN for the CIFAR10 dataset (see \cref{fig:figure_FAT_cifar10}) confirm that fast training techniques can be used with success also for adversarial training. Compared to the original FAT method, performing the training with a LR that follows the behavior of one of these above-discussed fast training policy can reduce the training time by around 2$\times$ (see pointer~1 - \cref{fig:figure_FAT_cifar10}), while obtaining a similar accuracy level (see pointer~2).

\subsection{Our Novel Contributions}
\begin{figure}[ht]
    \centering
    \includegraphics[width=\linewidth]{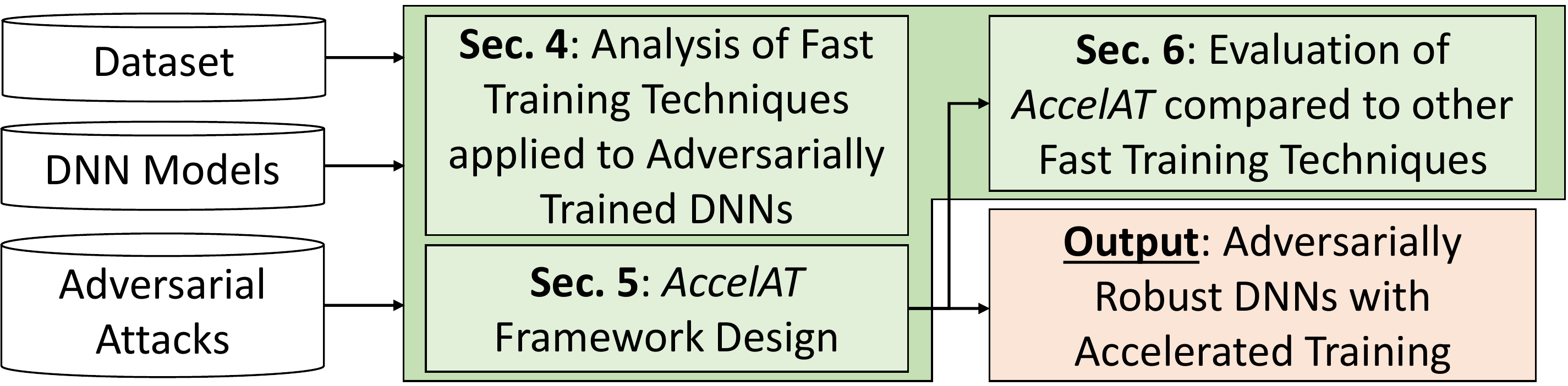}
    \caption{Overview of our novel contributions.}
    \label{fig:overview_novel_contributions}
\end{figure}

The main contributions of this paper are (see \cref{fig:overview_novel_contributions}):

\begin{itemize}[noitemsep,topsep=0pt]
    \item We analyze the prominent fast training techniques applied to adversarially trained DNNs, showing significant training time reduction in terms of training epochs~(\cref{sec:analysis}).
    \item We design a novel framework, \textit{AccelAT}, which automatically reduces the LR when the accuracy gradient starts decreasing, i.e., when the accuracy curve starts falling into a plateau region (\cref{sec:accelat}).
    \item The experimental results on multiple DNNs (ResNet, MobileNet) trained on CIFAR10 and CIFAR100 datasets with our \textit{AccelAT} framework obtain up to $8\%$ higher adversarial robustness against the most common attacks such as LinfPGD, Fast Gradient Sign Method (FGSM), and DeepFool (\cref{sec:results}).
\end{itemize}
\textbf{Open Source:} To facilitate the research and developments in this field, and for reproducible research, this is the AccelAT open-source framework: https://github.com/Nikfam/AccelAT.

Before proceeding to the main technical sections, we present an overview of adversarial attacks and defenses, and fast training policies for DNNs, in \cref{sec:backgroundadv} and \cref{sec:backgroundfast}, respectively, with a level of detail necessary to understand the rest of the paper.

\section{Overview of Adversarial Attacks and Defenses for DNNs}
\label{sec:backgroundadv}
Adversarial training~\cite{DBLP:journals/corr/abs-1810-09619} is a branch of ML that deals with creating robust models against adversarial attacks~\cite{DBLP:conf/cikm/RuanY021}, for instance, by augmenting the adversarial samples to the training dataset. For years, the training has only focused on achieving high accuracy. However, there exist malicious attacks that mine the algorithms correct behavior.  
If a DNN model is attacked, it will incorrectly execute its process, which can lead to severe consequences for safety-critical applications~\cite{DBLP:conf/iclr/KurakinGB17a}. For example, for cases where facial, voice, or fingerprint recognition is used to unlock certain services, an external attack can cause severe damage. Thus, it is desired that DNN models are robust against such external attacks. However, it is known that complex DNN models are vulnerable to malicious attacks, which could make their accuracy drop from near $100\%$ to nearly $0\%$~\cite{DBLP:journals/corr/GoodfellowSS14}. To counter these attacks, we need to develop and deploy robust models that can maintain high accuracy in the presence of these malicious variations. 

Adversarial training leverages clean images as well as the noisy images (following a certain adversarial attack model) for training the DNN models, thereby enabling the trained DNN models to classify correctly even in the presence of an adversarial attack during the inference. However, this robustness is achieved at the expense of significantly longer training time, proportional to the amount of adversarial samples and attack models.

\begin{figure}[ht]
    \centering
    \includegraphics[width=\linewidth]{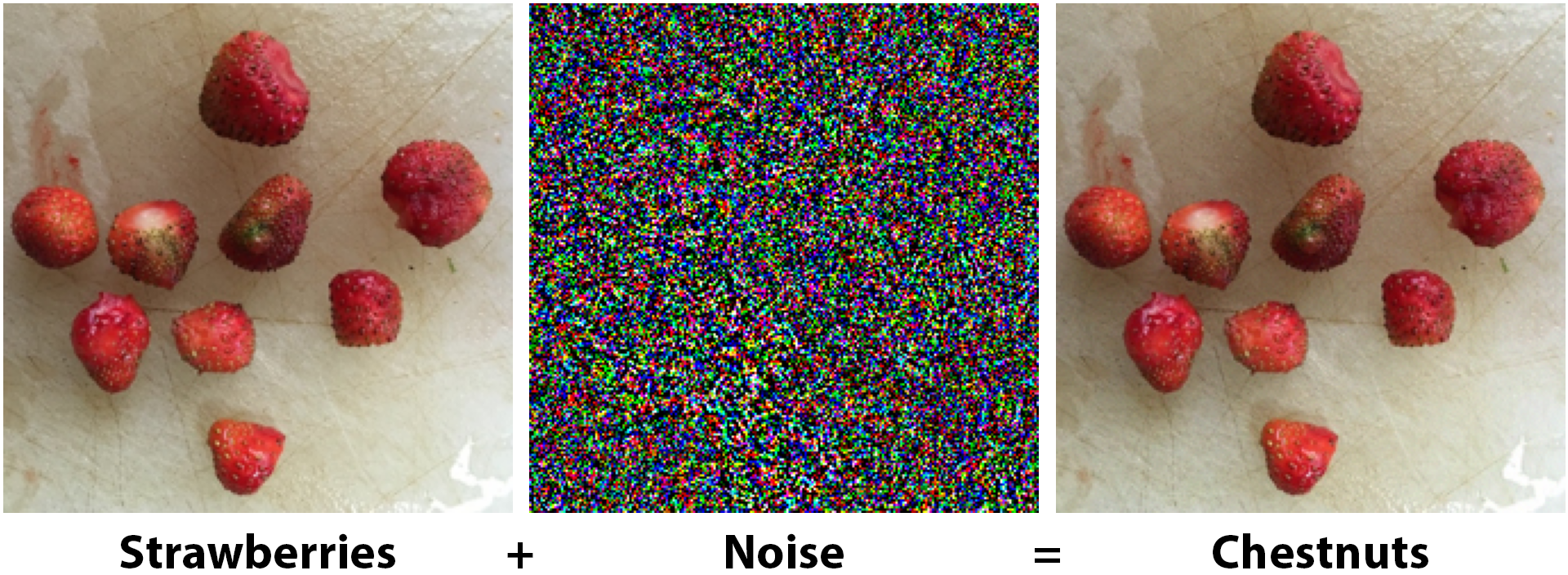}
    \caption{An example of adversarial attack, where strawberries are misclassified as chestnuts~\cite{DBLP:journals/corr/GoodfellowSS14}.}
    \label{fig:figure_strawberry}
\end{figure}

\subsection{Adversarial Examples}
The basic foundation of an adversarial attack algorithm is to create imperceptibly-modified examples~\cite{DBLP:journals/corr/GoodfellowSS14} that mislead the DNN model. On the other hand, if the examples were modified with a random logic, the problem would not arise since the model would fail, but a human being would easily recognize and detect the modifications. However, some examples could be modified to mislead a DNN model through a malicious attack without any evident variation perceived by the human eye. For example, as shown in \cref{fig:figure_strawberry}, the two images (original and modified by attack) are identical to the human eye. However, every single pixel has been modified according to the noise seen in the middle~\cite{DBLP:journals/corr/abs-2112-00639}. 

The consequence is that DNN models can be attacked without obvious external signs. In reality, with certain types of attacks, it is possible to generate modified images indistinguishable from a human being, which the DNN model can still correctly recognize. The latter case is not a critical problem, as they would be images discarded by a human. In this work, we will focus on the misclassification of DNN models.


\subsection{Adversarial Attacks}
A DNN model can learn to recognize images, but in an entirely different way from how humans do it. Therefore, various adversarial attacks algorithms have been proposed~\cite{DBLP:journals/csr/PitropakisPGAL19}. For example, some attacks are based on changing a single pixel~\cite{DBLP:journals/tec/SuVS19}, others on certain image features, but the most common approach consists of calculating the division line, called decision boundary, that distinguishes one class from another.

\begin{figure}[ht]
\centering
  \begin{subfigure}[b]{0.32\linewidth}
    \includegraphics[width=\textwidth]{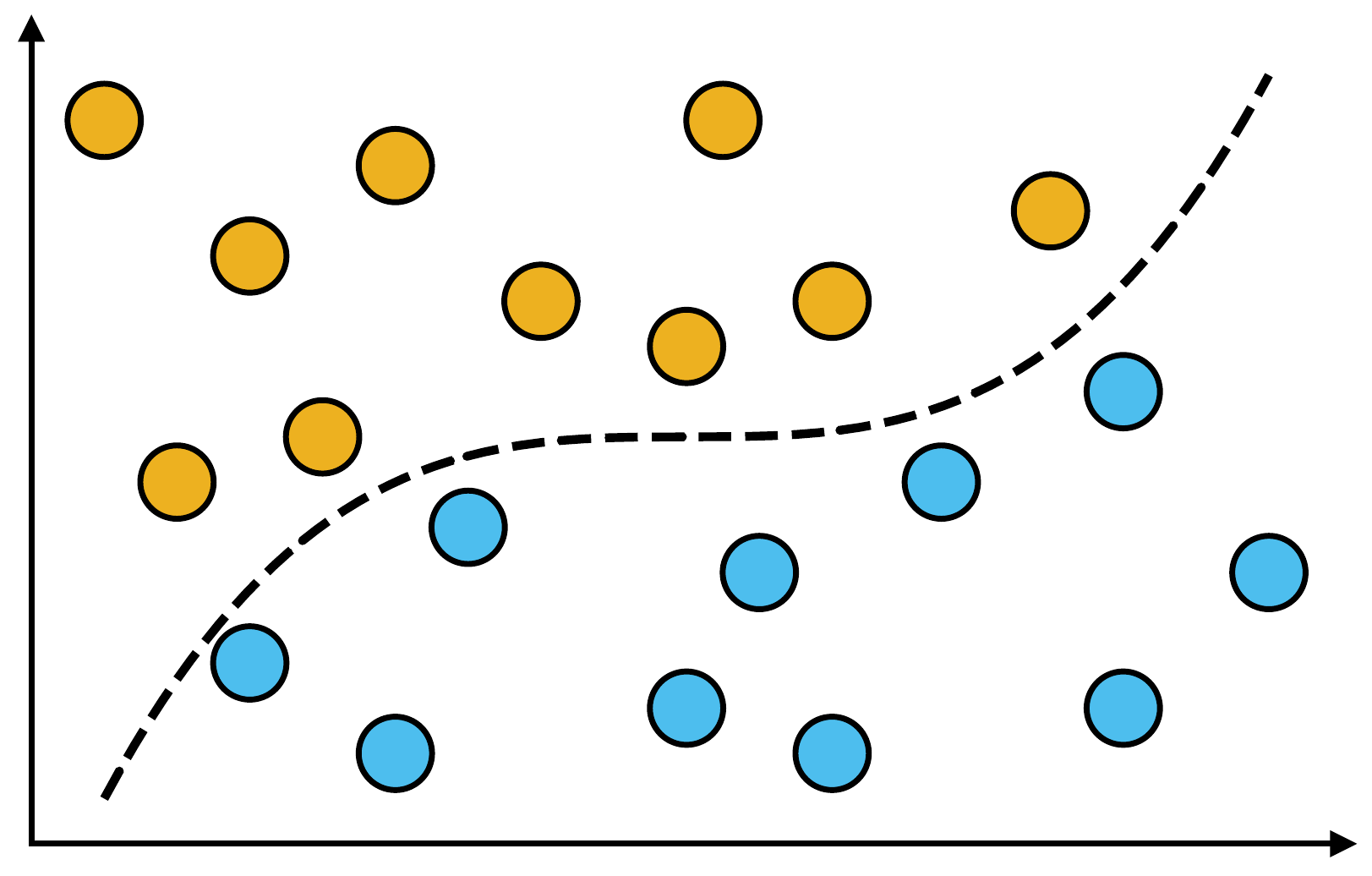}
    \caption{Decision line.}
    \label{fig:figure_line}
  \end{subfigure}
 \hfill
  \begin{subfigure}[b]{0.32\linewidth}
    \includegraphics[width=\textwidth]{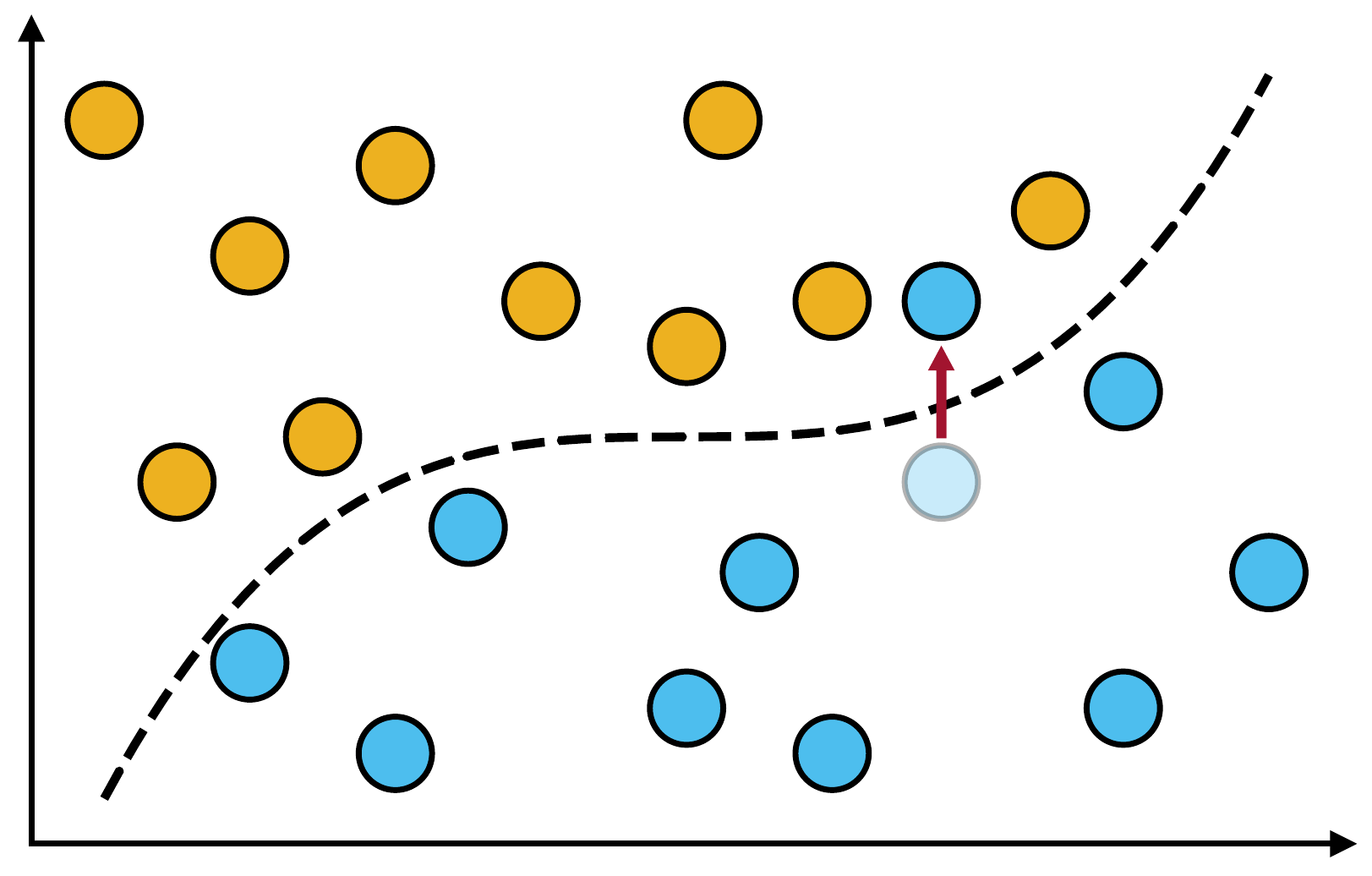}
    \caption{Pixels moving.}
    \label{fig:figure_line_cross}
  \end{subfigure}
 \hfill
  \begin{subfigure}[b]{0.32\linewidth}
    \includegraphics[width=\textwidth]{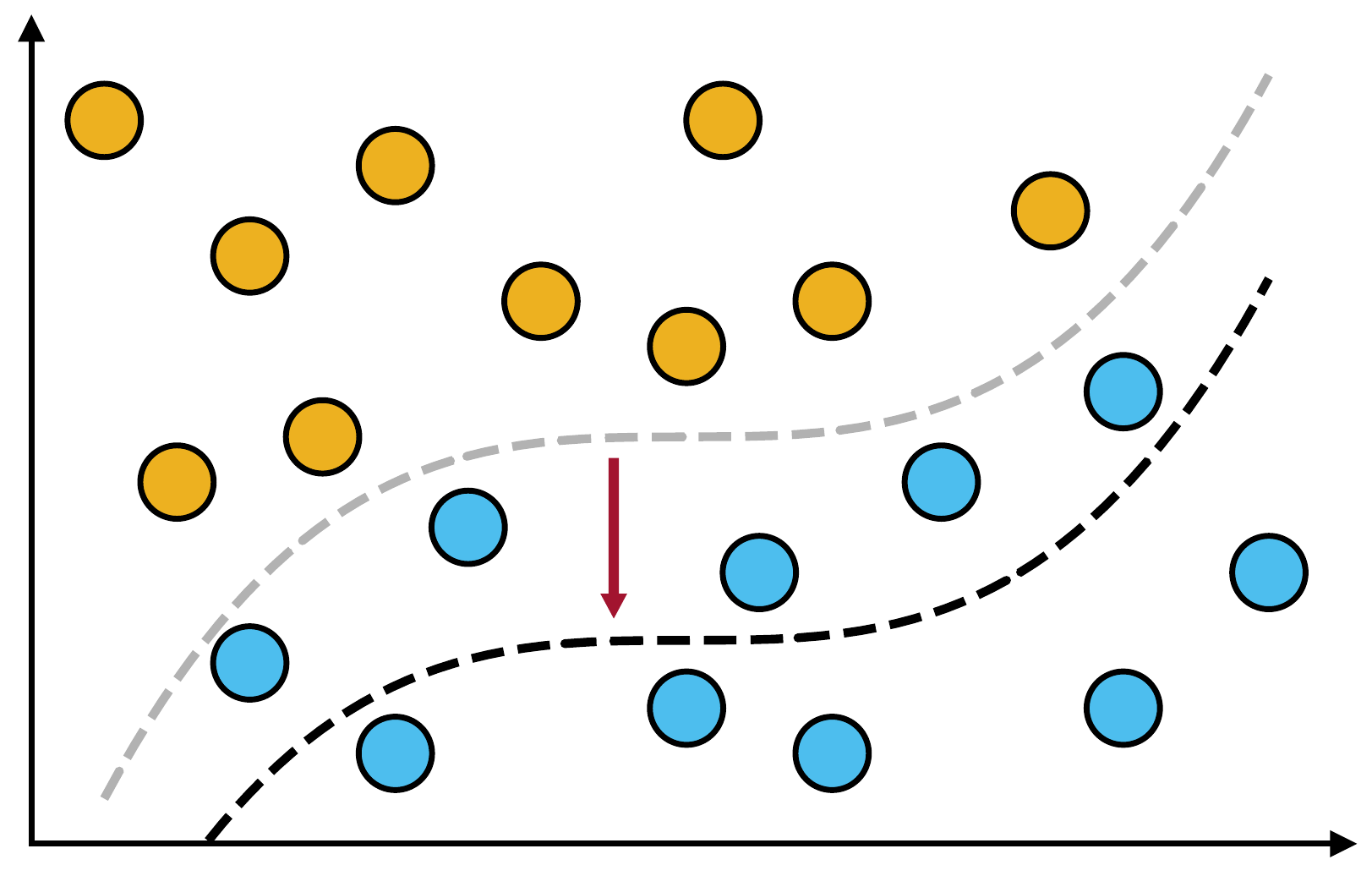}
    \caption{Line moving.}
    \label{fig:figure_line_move}
  \end{subfigure}
  \caption{Two ways of fooling a classifier.}
\end{figure}

In a simple case with two classes, the decision boundary looks like as in \cref{fig:figure_line}. A targeted attack would perturb all the borderline examples. For image classification applications, the attack would vary the last layer features, closest to the line edge, just enough to make them cross the line (\cref{fig:figure_line_cross}). In this way, the classification is completely distorted without actually changing the image much overall. This process deals with two almost identical images, which are instead classified in different ways. In some cases, the pixels do not change the position. On the contrary, the attack moves the separation line that distinguishes the classes (\cref{fig:figure_line_move}), thus leading to a misclassification due to the shifted decision boundary.

\subsection{White-Box Attacks}
There are mainly two categories of attacks, namely white-box and black-box~\cite{DBLP:conf/ccs/PapernotMGJCS17}. In this work, we focus on white-box attacks, which are the most accessible and powerful type of attacks to perform considering that the adversary has more knowledge about the system~\cite{DBLP:conf/iclr/MadryMSTV18}. While a black-box attack has access only to the inputs and outputs, a white-box attack also leverages the knowledge of the internal structure of the model to be attacked. Therefore, the attack is more specific and powerful, and the caused damage increases.

\subsection{Adversarial Training}
There exist various techniques to train robust models, such as data augmentation, second model control, and hyperparameters fine-tuning~\cite{DBLP:journals/csr/PitropakisPGAL19, DBLP:journals/tnn/YuanHZL19}.

The most common technique to counter attacks is data augmentation, that is, the DNNs are trained not only on correct images, but also on already attacked images or adversarial samples generated based on given attack models~\cite{DBLP:conf/nips/SchmidtSTTM18}. This procedure significantly increases the DNN accuracy against attacks~\cite{DBLP:journals/nn/EsfandiariBEVES21, DBLP:conf/iclr/ShafahiSZGSJG20}. However, as a drawback, the accuracy of clean images decreases, and often falls below the required accuracy level~\cite{DBLP:conf/nips/GuiWYYW019}.


The second model control uses two DNN models. This technique examines the main DNN and its internal characteristics to predict whether the analyzed example is adversarial or not~\cite{DBLP:conf/iclr/MetzenGFB17, DBLP:conf/cvpr/Moosavi-Dezfooli17}. In practice, this technique uses an “external guard” logic that controls the whole process to verify its effective operation. However, the study of its effectiveness and complexity is still immature.

Moreover, the performance of the adversarial training can be improved by changing the values of the hyperparameters, such as weight decay, batch size, or LR~\cite{DBLP:conf/iclr/PangYDSZ21,DBLP:journals/corr/abs-2010-03593}.

\subsection{Adversarial Libraries}
There are various libraries to implement the adversarial attacks~\cite{DBLP:journals/corr/GoodfellowPM16,DBLP:journals/corr/abs-2001-05574}. Foolbox~\cite{DBLP:journals/corr/RauberBB17} is used in this work due to its good documentation, functionality, and support for TensorFlow~\cite{DBLP:journals/corr/AbadiABBCCCDDDG16} and Pytorch~\cite{DBLP:conf/nips/PaszkeGMLBCKLGA19} packages.

\section{Overview of Fast Training Policies for DNNs}
\label{sec:backgroundfast}
We need fast training to meet an ever-increasing demand for large databases to be managed in real-time. For example, popular websites like Google, YouTube, and Facebook need to manage constant incoming data streams, training the models as quickly as possible. On the other hand, treating a large amount of data consumes a large amount of power, and is often out of the hand of small-scale organizations where training resources are limited. Therefore, accelerating this process allows obtaining advantages in terms of resources, time and energy.
With current processors, such as CPUs and GPUs, the computation times for the complete training can last from days to several weeks for large-sized datasets, and a few hours to days for medium-sized datasets. Hence, a significant reduction in the training time is highly desirable. There are various ways to speed up:
\begin{itemize}[noitemsep,topsep=0pt]
  \item Specialized techniques for certain types of DNN models, like Adam~\cite{DBLP:journals/corr/KingmaB14}, Ada-Boundary~\cite{song2019adaboundary}, or Super-Convergence~\cite{DBLP:journals/corr/abs-1708-07120}.
  \item Generic optimizations, like hyperparameters tuning~\cite{DBLP:journals/corr/abs-1803-09820}, which apply to nearly every DNN model.
\end{itemize}


\subsection{Fast Training Techniques}
Generic techniques mainly include the changes to hyperparameters~\cite{DBLP:journals/corr/abs-1803-09820} and, more specifically, to the LR~\cite{DBLP:conf/wacv/Smith17}, since variable LR values can give better results than constant values. Among the various state-of-the-art fast training methodologies proposed in the literature, the most advanced in this regard are the following:
\begin{itemize}[noitemsep,topsep=0pt]
  \item One cycle policy~\cite{DBLP:journals/corr/abs-1803-09820};
  \item Cyclical policy~\cite{DBLP:conf/wacv/Smith17};
  \item Warm restarts~\cite{DBLP:conf/iclr/LoshchilovH17, DBLP:conf/tencon/MishraS19}.
\end{itemize}
Before applying each of these techniques, it is necessary to find the best LR to use during the training, through the \textit{LR finder} technique.

\subsection{Learning Rate Finder}
\begin{figure}[ht]
    \centering
    \includegraphics[width=1\linewidth]{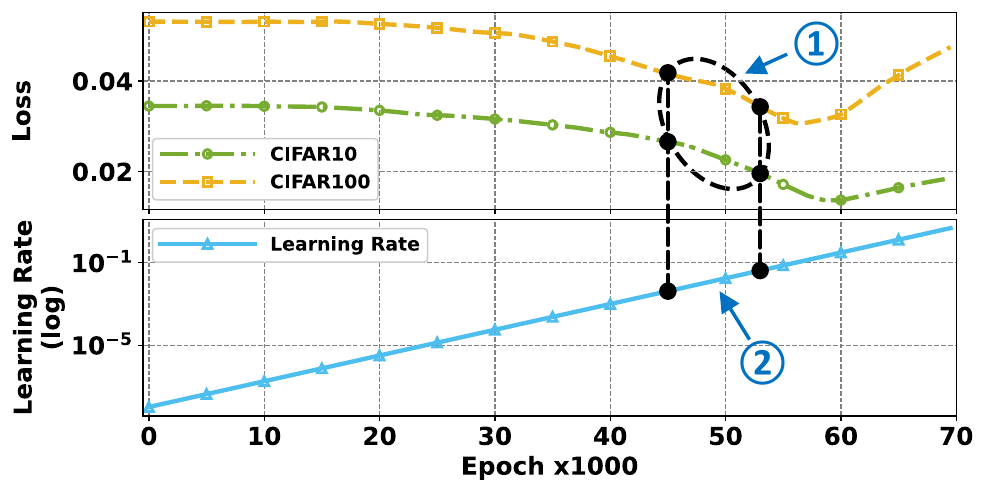}
    \caption{LR finder executed on the ResNet-50 model 
    for the CIFAR10 and CIFAR100 datasets.}
    \label{fig:figure_lr_finder}
\end{figure}

The simplest method to find the correct LR value is to change it exponentially, from small to large values, during reasonably long training. An efficient choice is to vary the LR by at least ten orders of magnitude throughout the training. Theoretically, if all the various parameters have been normalized, the LR will often lay between $0.001$ and $10$. For this reason, it is preferable to fully cover this range during the training. Once the test training is finished, it will be enough to look at the accuracy and error graph to find the maximum recommended LR~\cite{DBLP:conf/wacv/Smith17}.
As shown in \cref{fig:figure_lr_finder}, for small LR values, the accuracy and the error practically do not vary. However, in a certain range of LR values, the error decreases up to a minimum, after which it diverges for large LR values. Hence, the desirable LR values are those in which the error decreases from the initial plateau to the minimum point, beyond which the divergence begins. Therefore, using a maximum LR value of about one order of magnitude lower than the minimum error point is advisable to be distant enough to avoid the divergence region (see pointer~1 - \cref{fig:figure_lr_finder}). In summary, the LR should range from the value in which the error slope starts decreasing, until one order of magnitude less than the point in which the error curve exhibits the minimum (see pointer~2).

\subsection{One Cycle policy}
The one cycle policy~\cite{DBLP:journals/corr/abs-1803-09820} is based on varying the LR and other hyperparameters during the training process to obtain fast training. As the name implies, the basic idea is to apply a single cycle to these hyperparameters throughout the training. Since the one cycle policy is a regularization technique, other types of normalization affecting hyperparameters must be reduced to avoid interference.

\begin{figure}[ht]
    \raggedright
    \includegraphics[width=1\linewidth]{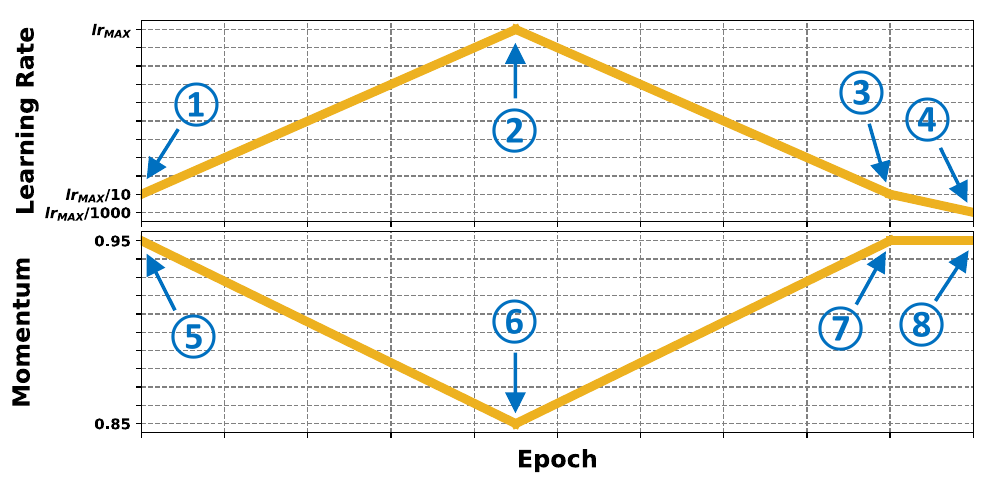}
    \caption{One cycle policy.}
    \label{fig:figure_1cycle}
\end{figure}

After finding the maximum LR through the LR finder (\cref{fig:figure_lr_finder}), an initial value equal to $1/10$ of the maximum is set (see pointer~1 - \cref{fig:figure_1cycle}). Then, the LR assumes the shape of a triangular cycle for about $90\%$ of the total training, i.e., $90\%$ of the total epochs (see \cref{fig:figure_1cycle}), first increasing from the initial value up to maximum (see pointer~2), and then decreasing again to $1/10$ of the maximum (see pointer~3). In the last few epochs, equal to about $10\%$ of the total epochs, the LR rapidly decreases to $1/1000$ of the maximum LR (see pointer~4). Properly setting the duration of the last part of the training is extremely important, since a longer duration would lead to overfitting, while a shorter duration would lead to low accuracy.

The one cycle policy is also applied to the momentum with an opposite shape (\cref{fig:figure_1cycle}). In this way, the regularization carried out on the LR is not dampened by the momentum, but on the contrary, it is strengthened. There is a maximum recommended momentum value of $0.95$ (see pointer~5), while the minimum should be $0.85$ (see pointer~6). In the final part of the training, while the LR decreases rapidly, the momentum remains fixed at the maximum value of $0.95$ (see pointers 7 and 8).

\subsection{Cyclical policy}
\begin{figure}[ht]
    \centering
    \includegraphics[width=0.9\linewidth]{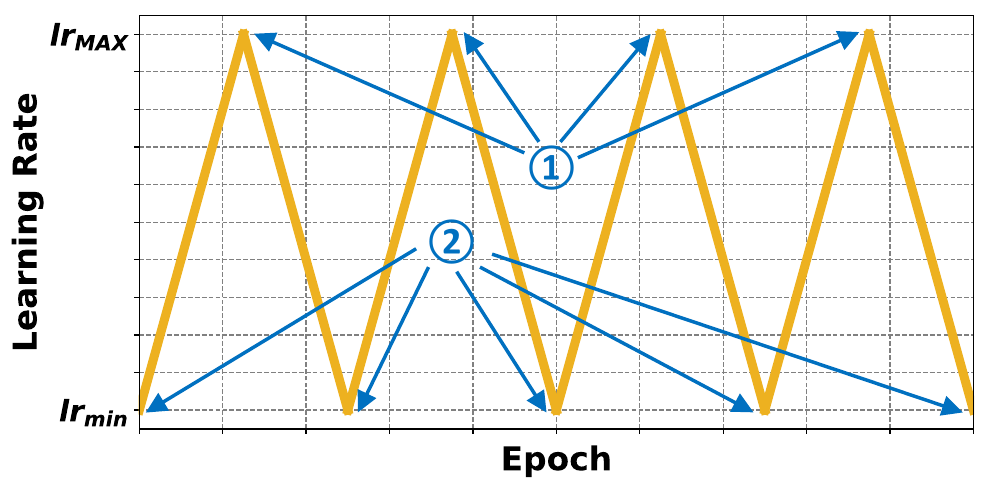}
    \caption{A triangular cyclical policy~\cite{DBLP:conf/wacv/Smith17}.}
    \label{fig:figure_cyclical_linear}
\end{figure}

\begin{figure}[ht]
    \centering
    \includegraphics[width=0.9\linewidth]{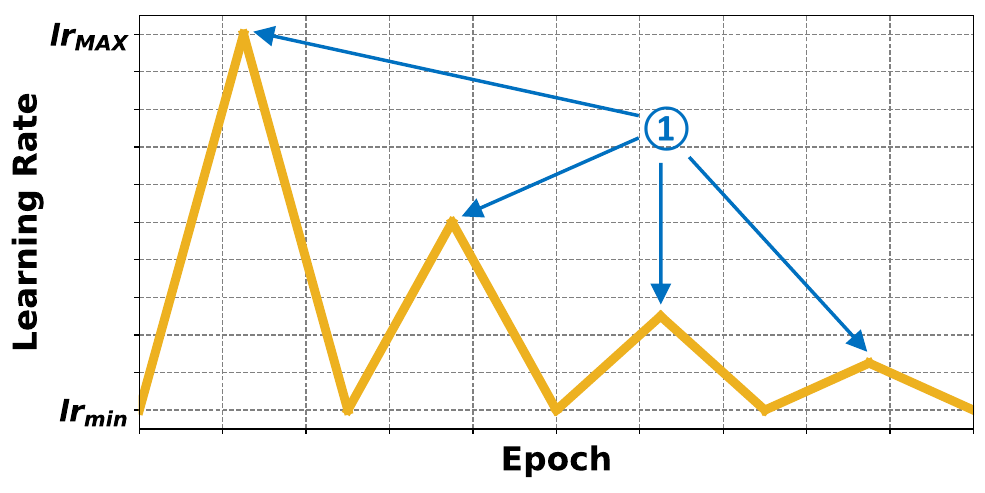}
    \caption{Cyclical policy with fixed lower boundary.}
    \label{fig:figure_cyclical_decreasing}
\end{figure}

The cyclical policy~\cite{DBLP:conf/wacv/Smith17}, shown in \cref{fig:figure_cyclical_linear,fig:figure_cyclical_decreasing}, is similar to the one cycle policy, with the difference that the cycle is repeated several times, constantly oscillating between the same maximum (see pointer~1 - \cref{fig:figure_cyclical_linear}) and minimum (see pointer~2) values. This policy can be helpful if the training process of the DNN model exhibit many local minimum points, since using a cyclical LR allows the training to seek deeper minimums and achieve higher accuracy.

The length of every single cycle is calculated as a multiple of an epoch. It is recommended to use cycle length values between $4$ and $20$ times an epoch to obtain optimal results. However, it is advisable to perform training with at least $3$ - $5$ cycles to obtain an evident improvement over a constant LR. Increasing the number of cycles too much would eliminate the cycle's usefulness, because the training would not have time to adapt to the variation of the LR.

\begin{figure}[ht]
    \centering
    \includegraphics[width=1\linewidth]{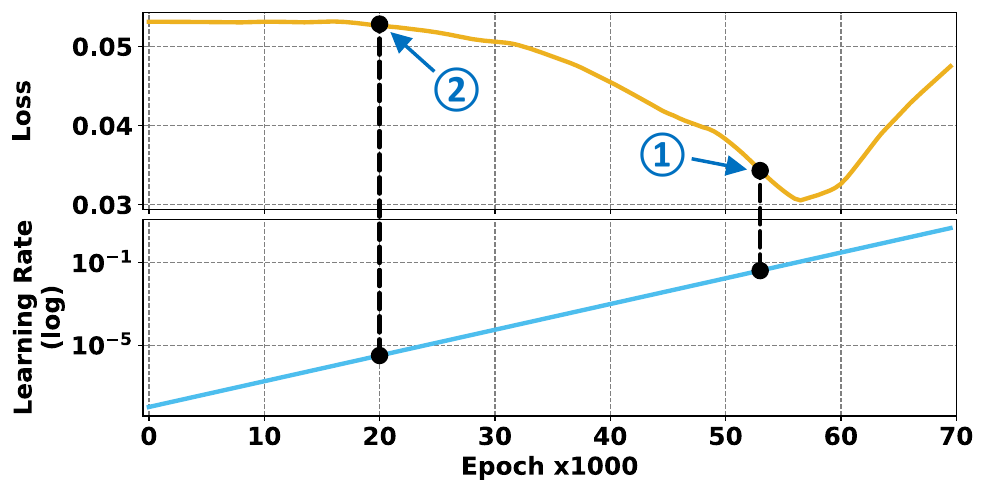}
    \caption{LR boundary on the loss plot for the cyclical policy~\cite{DBLP:conf/wacv/Smith17}.}
    \label{fig:figure_cyclical_boundary}
\end{figure}

The maximum and minimum values of the LR to adopt in the cycle must be chosen carefully  (\cref{fig:figure_cyclical_boundary}), since the success of the training depends on them. In both cases, it is necessary to use the graph produced by the LR finder (\cref{fig:figure_lr_finder}), which must be run before the final training. The maximum LR is found precisely as for the one cycle policy, i.e., $1/10$ of the minimum point of the loss that corresponds to the limit (see pointer~1 - \cref{fig:figure_cyclical_boundary}). On the other hand, the minimum LR is set to a value in the loss descent zone from the initial plateau onwards (see pointer~2 - \cref{fig:figure_cyclical_boundary}).

In some other cases, the cycles are repeated with the same length, but the maximum LR value decreases (see pointer~1 - \cref{fig:figure_cyclical_decreasing}) to search deeper in the local minima, such as the decreasing triangular cycles of \cref{fig:figure_cyclical_decreasing}.

\subsection{Warm restarts}
\begin{figure}[ht]
    \centering
    \includegraphics[width=0.9\linewidth]{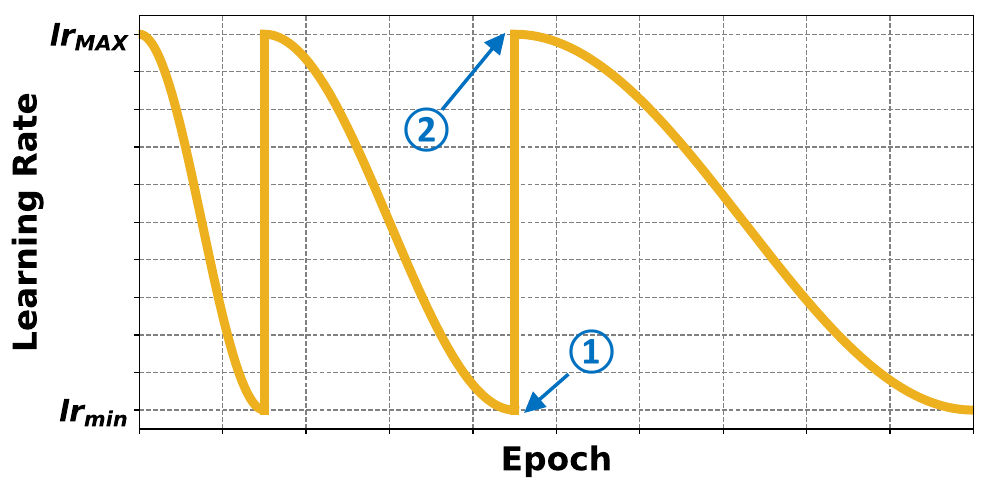}
    \caption{Accordion-like sinusoidal warm restarts~\cite{DBLP:conf/iclr/LoshchilovH17}.}
    \label{fig:figure_warm_sinusoidal}
\end{figure}

\begin{figure}[ht]
    \centering
    \includegraphics[width=0.9\linewidth]{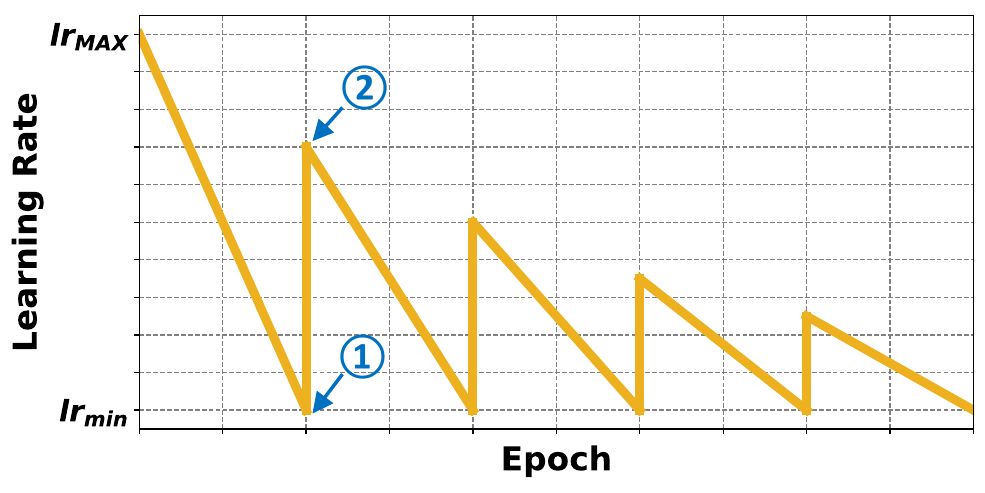}
    \caption{Linear decreasing warm restarts~\cite{DBLP:conf/tencon/MishraS19}.}
    \label{fig:figure_warm_linear}
\end{figure}

The warm restart methods~\cite{DBLP:conf/iclr/LoshchilovH17}~\cite{DBLP:conf/tencon/MishraS19} are also based on a cyclical policy, but as the term says, there are sudden restarts from the minimum (see pointer~1 - \cref{fig:figure_warm_sinusoidal}, pointer~1 - \cref{fig:figure_warm_linear}) to the maximum LR value (see pointer~2 - \cref{fig:figure_warm_sinusoidal}, pointer~2 - \cref{fig:figure_warm_linear}). This phenomenon leads to instantaneously restart a new long descent, aiming at finding deeper minima. The warm restarts are always performed on the LR and can assume various shapes, such as:
\begin{itemize}[noitemsep,topsep=0pt]
  \item Sinusoidal (\cref{fig:figure_warm_sinusoidal});
  \item Linear (\cref{fig:figure_warm_linear});
  \item Trapezoidal.
\end{itemize}

The warm restarts can have multipliers that make the progress accordion-like during the training (\cref{fig:figure_warm_sinusoidal}), or the restarts can be at different gradually decreasing values (\cref{fig:figure_warm_linear}).

\subsection{Other Fast Training Methods}
Changing the shapes of the hyperparameters or mixing the above-discussed techniques, it possible to obtain new training policies for the LR, which might be more effective than the originals methods.

\section{Analysis: Fast Training Techniques Applied to Adversarial Training Methods}
\label{sec:analysis}
Various methodologies have recently been proposed to accelerate the adversatial training~\cite{DBLP:conf/nips/ZhangZLZ019, DBLP:conf/icml/ZhangYJXGJ19}. The Free Adversarial Training (FAT)~\cite{DBLP:conf/nips/ShafahiNG0DSDTG19} is chosen as the baseline method for the experiments in this section.

The $m$ parameter, also called \textit{Free-m}, is a key parameter of the FAT algorithm, since it allows to repeat the perturbation several times for every single minibatch~\cite{DBLP:journals/corr/abs-1712-02029}. With this terminology, a traditional training is obtained by keeping $m = 1$. The parameter $\epsilon$ indicates the adversarial perturbation. A too large $\epsilon$ would make the perturbations so high that the images would be recognized as crafted even by the human eye.

Building on top of this, there have been concurrent works such as Fast is Better than Free~\cite{DBLP:conf/iclr/WongRK20} and subsequently, also GradAlign~\cite{DBLP:conf/nips/AndriushchenkoF20}, which have demonstrated the reliability in using the FGSM for speeding up with the proper precautions. In this work, we mainly focus, instead of on the type of training, on the hyperparameters. Therefore, we tested the feasibility of \textit{AccelAT} also with the FGSM method.

\subsection{Original FAT Results}
\begin{figure}[ht]
    \centering
    \includegraphics[width=1\linewidth]{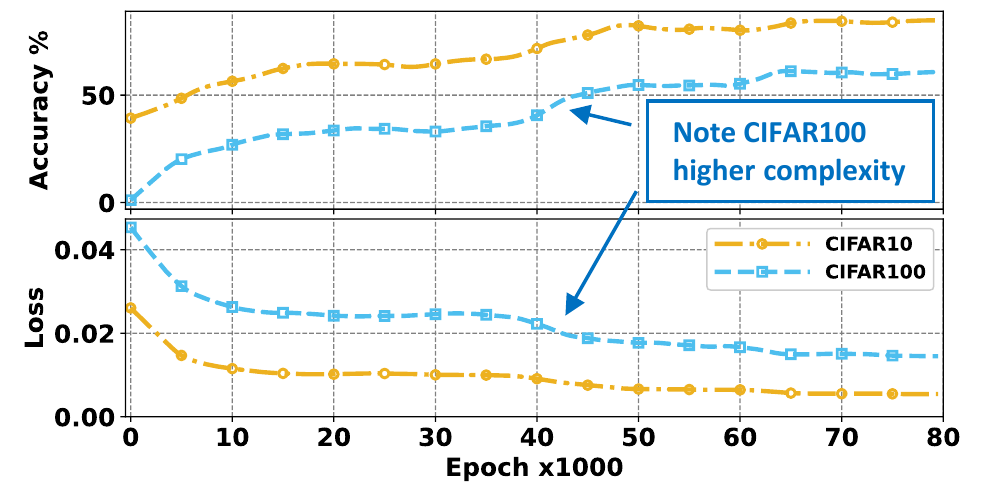}
    \caption{Original FAT~\cite{DBLP:conf/nips/ShafahiNG0DSDTG19} accuracy and loss on natural images.}
    \label{fig:figure_FAT_original}
\end{figure}

A first analysis has been conducted by reproducing the original FAT method, applied to the ResNet-50 model~\cite{DBLP:conf/cvpr/HeZRS16} on CIFAR10~\cite{cifar10, Krizhevsky09learningmultiple}  and CIFAR100~\cite{cifar100, Krizhevsky09learningmultiple} datasets, under the projected gradient descent (PGD) attack~\cite{DBLP:conf/icml/LiuY09} with constant $\epsilon$ equal to $8.0$. 
\cref{fig:figure_FAT_original} shows the training results in terms of accuracy and loss, obtained for both the CIFAR10 and the CIFAR100 datasets on natural images. As expected, the CIFAR100 accuracy is lower than the CIFAR10 accuracy due to the higher complexity.
Final accuracy results:
\begin{itemize}[noitemsep,topsep=0pt]
  \item CIFAR10 \textrightarrow{} accuracy: $84.34\%$ - loss: $0.00562$;
  \item CIFAR100 \textrightarrow{} accuracy: $59.89\%$ - loss: $0.01459$.
\end{itemize}

\subsection{Hyperparameters Setup}
\begin{figure}[ht]
    \raggedright
    \includegraphics[width=0.90\linewidth]{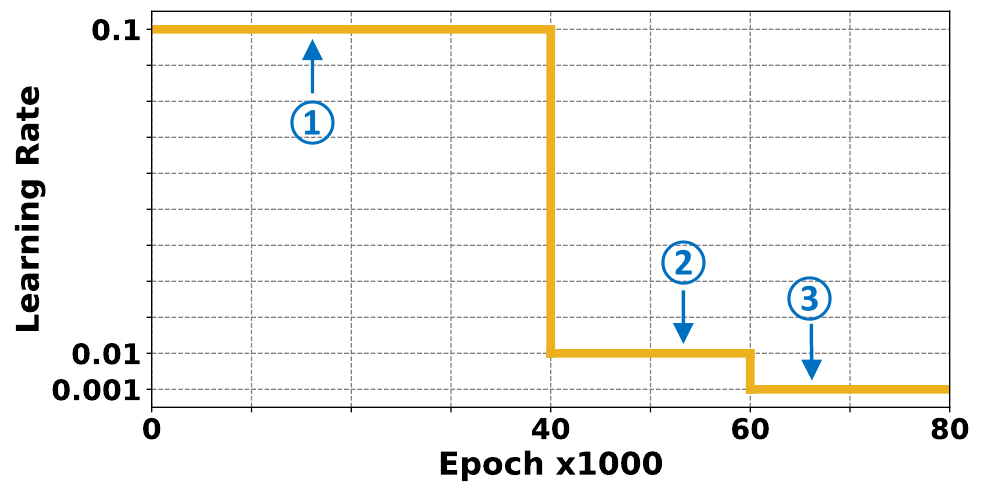}
    \caption{FAT’s 3-steps LR~\cite{DBLP:conf/nips/ShafahiNG0DSDTG19}.}
    \label{fig:figure_steps}
\end{figure}

The original FAT method applies a LR that has a 3-step function (\cref{fig:figure_steps}) with the following behavior:
\begin{itemize}[noitemsep,topsep=0pt]
  \item Epochs = $[0 : 40000]$ \textrightarrow{} LR = $0.1$ (see pointer~1 - \cref{fig:figure_steps});
  \item Epochs = $[40000 : 60000]$ \textrightarrow{} LR = $0.01$ (see pointer~2);
  \item Epochs = $[60000 : 80000]$ \textrightarrow{} LR = $0.001$ (see pointer~3).
\end{itemize}
After using the LR finder (\cref{fig:figure_lr_finder}), the maximum LR value results to be:
\begin{itemize}[noitemsep,topsep=0pt]
  \item CIFAR10 \textrightarrow{} maximum LR = $0.15$;
  \item CIFAR100 \textrightarrow{} maximum LR = $0.12$.
\end{itemize}
Therefore, for the experiments applying the fast training techniques on the FAT method, a maximum LR higher than that of the original FAT is used.

The momentum values are set to:
\begin{itemize}[noitemsep,topsep=0pt]
  \item One cycle \textrightarrow{} momentum = $0.85$ – $0.95$ (\cref{fig:figure_1cycle});
  \item Constant \textrightarrow{} momentum = $0.90$.
\end{itemize}
The one cycle momentum is used only for the one cycle policy~\cite{DBLP:journals/corr/abs-1803-09820}. Instead, for the other techniques, the momentum is fixed to the original FAT constant value.

Based on the regularization criteria, the value of the weight decay has been set to $0.0002$. A batch size of $128$ has been set due to the computational limits of the calculator. The remaining FAT parameters relative to adversarial training have not been changed, since the aim is not to obtain a more robust model, but to accelerate the training, while achieving the same robustness.

\subsection{Super FAT results}
\begin{figure}[ht]
    \centering
    \includegraphics[width=1\linewidth]{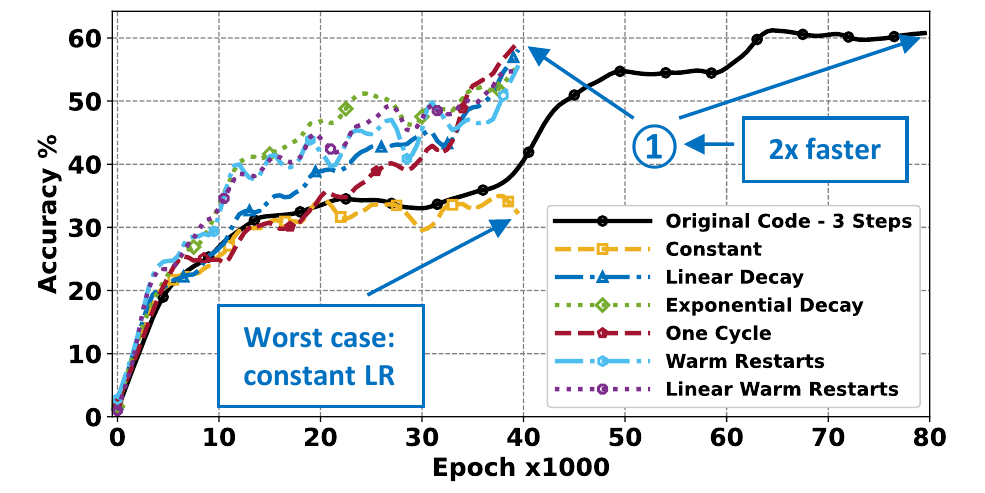}
    \caption{FAT Training of ResNet-20 on the \textit{CIFAR100} natural images dataset with different LR techniques.}
    \label{fig:figure_FAT_cifar100}
\end{figure}

The results of different training policies applied to the FAT, compared to the original FAT, for the CIFAR10 and CIFAR100 natural images datasets, are shown in \cref{fig:figure_FAT_cifar10} and \cref{fig:figure_FAT_cifar100}, respectively.
The experiments are performed with various values of training epochs to show the differences between the algorithms. With our settings (i.e., execution on the Tesla K40c GPU), $10000$ epochs are executed in about $5$ hours. Therefore, the original FAT training lasts for about $40$ hours. For this reason, halving the number of epochs allows to execute the experiments faster.

From these results, it can be noticed that the FAT algorithm with a 3-steps LR is already optimized w.r.t. the original adversarial training~\cite{DBLP:conf/iclr/MadryMSTV18}, but with more advanced techniques, the same results can be achieved even in about half the time (see pointer~1 - \cref{fig:figure_FAT_cifar100}). Thus, the fine-tuning of the hyperparameters is essential and can lead to super-convergence in standard training and adversarial training without affecting the result and robustness of the DNN model itself.

\section{Our \textit{AccelAT} Framework}
\label{sec:accelat}

The idea behind \textit{AccelAT} is to avoid or reduce the model setup time. With the existing fast training techniques, we can obtain good results, but often it takes too much time to set up all parameters and several attempts before finding the best compromise. These dead times are not considered in the training time, but they are still a considerable part of a programmer's work. With \textit{AccelAT}, it is possible to have more freedom of choice since it will be the model itself during the training that will understand when to intervene on the LR for more efficient learning. As we will see, this leads to similar or better results to existing fast training techniques based on the LR.

The functionality of our \textit{AccelAT} methodology is described in \cref{alg:algorithm_1}. The LR value, initially equal to the maximum possible value obtained with the LR finder, decreases as plateau zones are found in the learning curve. The LR decreases if the accuracy does not exceed a specific $\Delta_{acc}$ in a certain number of epochs.

\begin{algorithm}
\label{alg:accelat}
\textbf{Require:} Maximum learning rate \textit{lr\textsubscript{MAX}}, minimum learning rate \textit{lr\textsubscript{min}}, accuracy delta \textit{\textDelta\textsubscript{acc}}, percentage reduction \textit{p}, number of cycles of interest \textit{n}, accuracy \textit{acc}, previous average accuracy \textit{acc\textsubscript{pre}}
  \caption{\textit{AccelAT}}
  \begin{algorithmic} [1]
    \STATE $lr\gets lr_{MAX}$
    \FOR{$e$ in $epochs$}
      \IF{$(\overline{acc(e,e-n)} - acc_{pre}) < \Delta_{acc}$}
        \STATE $lr\gets lr \cdot p$
      \ENDIF
      \IF{$lr < lr_{min}$}
        \STATE $lr\gets lr_{min}$
      \ENDIF
      \STATE $acc_{pre}\gets \overline{acc(e,e-n)}$
    \ENDFOR
  \end{algorithmic}
  \label{alg:algorithm_1}
\end{algorithm}

\begin{figure}[ht]
    \centering
    \includegraphics[width=1\linewidth]{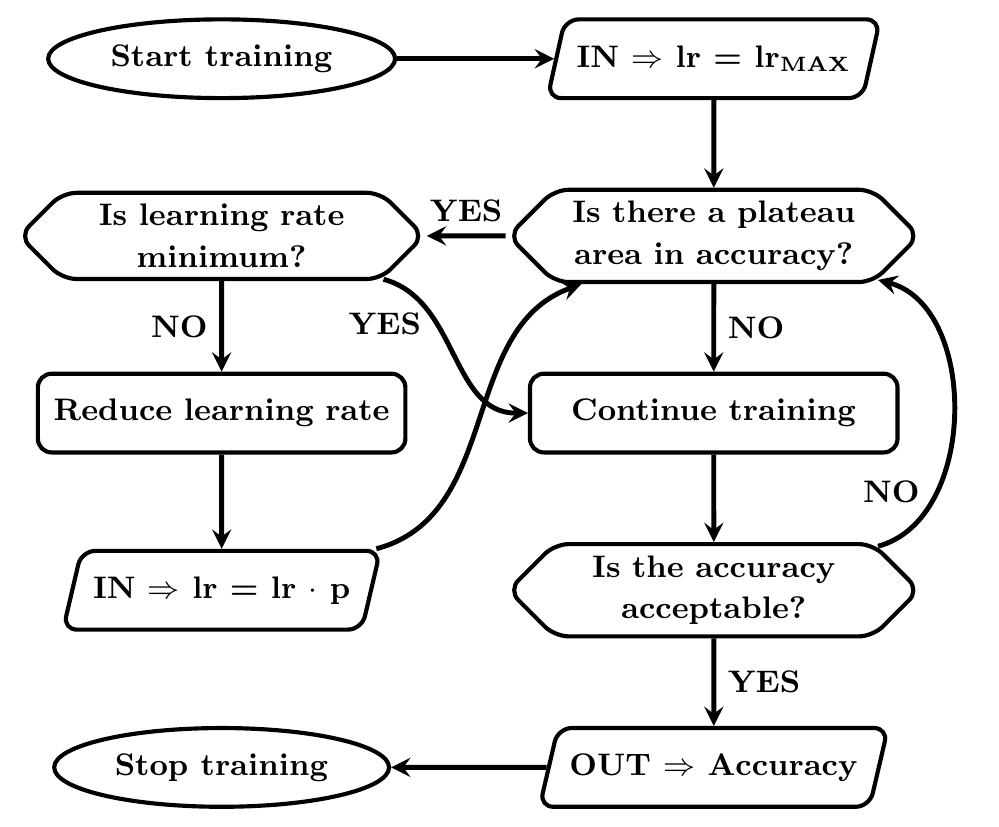}
    \caption{\textit{AccelAT} workflow.}
    \label{fig:figure_workflow}
\end{figure}

Inspired by existing fast training techniques that eliminate plateau areas while learning, in our \textit{AccelAT} framework the LR is varied based on the performance of the validation accuracy. First, we search for the maximum LR using the LR finder technique, after which we set it as the initial LR (line 1 - \cref{alg:algorithm_1}), to be decreased if the accuracy starts to show a plateau. Then, a simplified gradient can be used to change the LR based on accuracy progression. As indicated in lines 3-4, the LR is decreased by a percentage value $p$ if the accuracy in the last $n$ cycles has not increased by a certain value $\Delta_{acc}$, up to the minimum desired LR (see \cref{fig:figure_workflow}).

Step by step, the framework works as follows:
\begin{itemize}[noitemsep,topsep=0pt]
    \item With previous training, we find the maximum LR that can be used through the LR finder;
    \item The LR is set to the previously found maximum LR value;
    \item The training starts with the predetermined LR;
    \item The accuracy is calculated for the dataset under analysis;
    \item If the accuracy does not grow at a sufficient rate that is decided a priori, then the LR decreases by a fixed percentage;
    \item If, on the other hand, there are no plateau areas, then the training continues with the current LR;
    \item Once the LR reaches a predetermined minimum value, then the training continues with this fixed LR value;
    \item When the accuracy reaches an optimal value, or it is impossible to rise further, then the training stops.
\end{itemize}

For example, let us set a value (\textit{n}) of ten epochs to evaluate the accuracy, a delta (\textit{\textDelta\textsubscript{acc}}) equal to $1\%$ and let us assume to have found a value of 0.01 as the maximum LR, which we want to reduce by about $10\%$ (\textit{p}) each time there is a plateau area.
After that, the training is launched for 100 epochs. The accuracy increase rate is monitored for each epoch. For instance, in the first 40 epochs, the increase in accuracy is greater than or equal to $1\%$, compared to the last ten epochs. Hence, the LR remains fixed at 0.01.
At the 41st epoch, the accuracy has not increased by at least $1\%$ in the last ten epochs, and consequently, the LR is multiplied by 0.9, i.e., it is reduced by $10\%$, and we obtain an LR value equal to 0.009.
Afterward, the training is resumed and the accuracy starts to rise again. Towards the 70th epoch, it again has a new plateau zone; consequently, the LR is reduced by another $10\%$.
Then, the training continues until the end of the 100 epochs. Once finished, the LR is lower than the maximum, allowing us to increase the accuracy by going deeper into the found local minimum. If we had kept a fixed LR after the first 40 epochs, the accuracy would not have increased, and the last 60 epochs would have been useless. With our \textit{AccelAT} framework, we have obtained an adaptive LR based on the specific training we are performing.

This type of LR policy does not have a fixed shape for every training and, therefore, cannot be plotted. Instead, its shape varies at run-time according to the model, dataset, and training parameters. In any case, it will assume a ladder shape.
Compared to other fast training techniques, the \textit{AccelAT} method, not having a fixed shape and calculating the gradient at run-time, is likely to slightly slow down the training. However, on long training with complex datasets, the \textit{AccelAT} allows obtaining an LR suited to the situation.

As will be demonstrated in the next section, our \textit{AccelAT} is more efficient than existing fast training techniques in certain types of training. Therefore, the best approach is to conduct preliminary analyses to determine which are good values of the parameters $p$, $n$, and $\Delta_{acc}$ to set for the training.

\section{Evaluating our \textit{AccelAT} Framework}
\label{sec:results}

\subsection{Experimental Setup}

\begin{figure}[ht]
    \centering
    \includegraphics[width=1\linewidth]{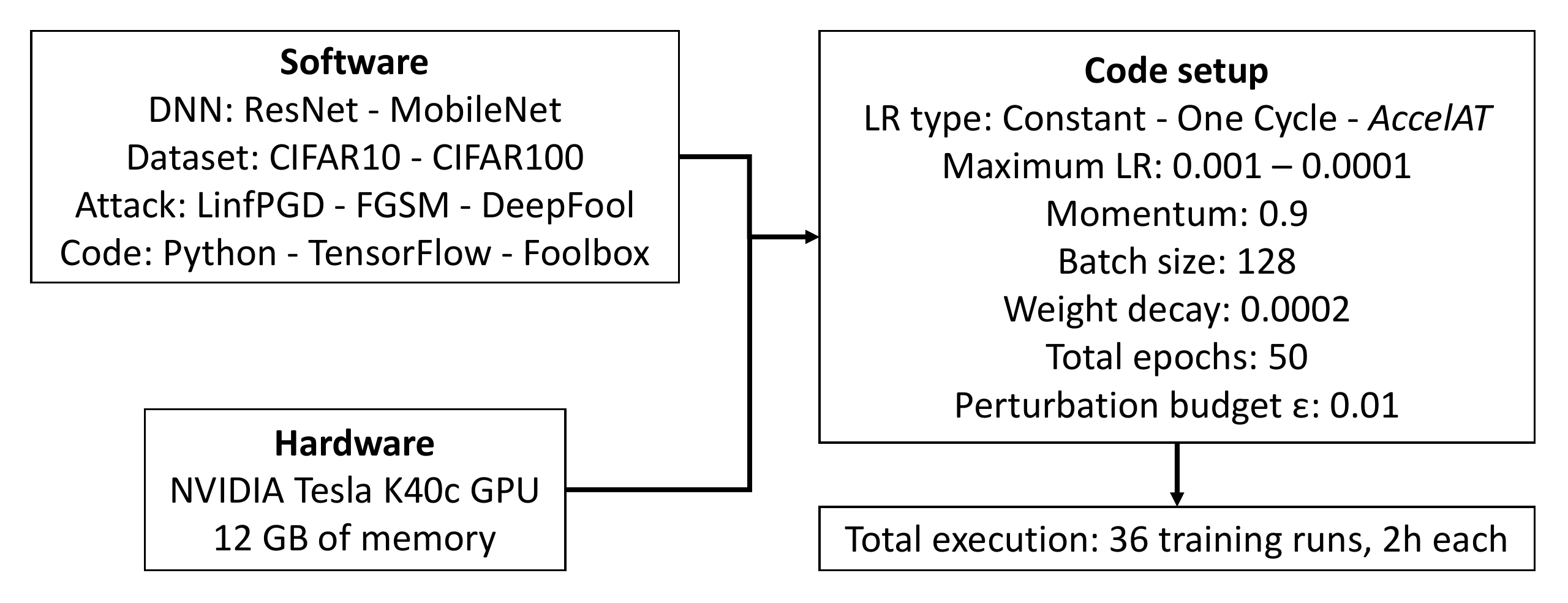}
    \caption{Experimental setup.}
    \label{fig:figure_setup}
\end{figure}

As shown in \cref{fig:figure_setup}, our experiments has been conducted using two popular DNNs (ResNet~\cite{DBLP:conf/cvpr/HeZRS16} and MobileNet~\cite{DBLP:journals/corr/HowardZCKWWAA17}), pre-trained with ImageNet, then trained for the CIFAR10~\cite{cifar10, Krizhevsky09learningmultiple} and CIFAR100~\cite{cifar100, Krizhevsky09learningmultiple} datasets, analyzed through the LinfPGD, FGSM, and DeepFool attacks described through the Foolbox library~\cite{DBLP:journals/corr/RauberBB17}. The code is written in Python~\cite{ml_raschka_2017, DBLP:journals/information/RaschkaPN20} using the TensorFlow 2.X library~\cite{DBLP:journals/corr/AbadiABBCCCDDDG16}, running on an NVIDIA Tesla K40c GPU with 12 GB of memory.

Each network/dataset/attack combination is tested with three types of learning policies, which are constant (the most simple one), one cycle (the best during FAT review), and \textit{AccelAT}. Therefore, we perform $36$ total training runs, taking around $2$ hours per run. We found maximum LR values between $0.001$ and $0.0001$ for all simulations using the LR finder. The other hyperparameters are kept at a fixed optimal value for our setup for all simulations:

\begin{itemize}[noitemsep,topsep=0pt]
  \item Momentum \textrightarrow{} $0.9$;
  \item Batch size \textrightarrow{} $128$;
  \item Weight decay \textrightarrow{} $0.0002$.
\end{itemize}

To avoid overfitting, we have noticed that the best choice is to use early stopping for the training. As also highlighted in these papers~\cite{DBLP:conf/icml/RiceWK20, DBLP:conf/icml/ZhangXH0CSK20}, early stopping can lead to more optimal solutions than the introduction of new normalizations. For this reason, we set our total epochs to $50$ to avoid overfitting. This choice does not conflict with our algorithm as acceptable accuracy values are obtained even with fewer epochs.
For the attack we used a perturbation budget $\varepsilon$ equal to 0.01 and the algorithm automatically iterates until it reaches an acceptable level to fool the DNN~\cite{DBLP:journals/corr/abs-1902-06705}.

\subsection{Experimental Results}

\begin{figure}[ht]
    \centering
    \includegraphics[width=1\linewidth]{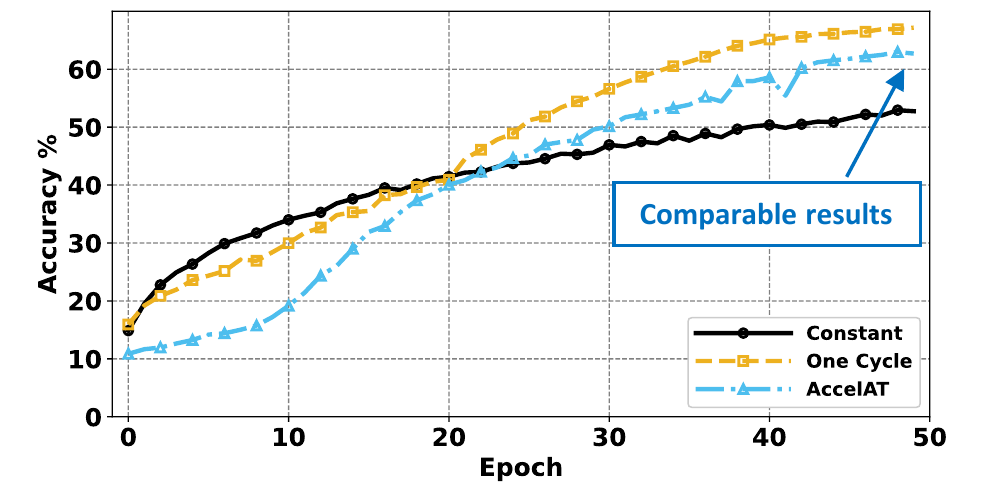}
    \caption{ResNET model trained on the CIFAR10 dataset, attacked with DeepFool - Adversarial training-set.}
    \label{fig:figure_res_deep_10}
\end{figure}

\begin{figure}[ht]
    \centering
    \includegraphics[width=1\linewidth]{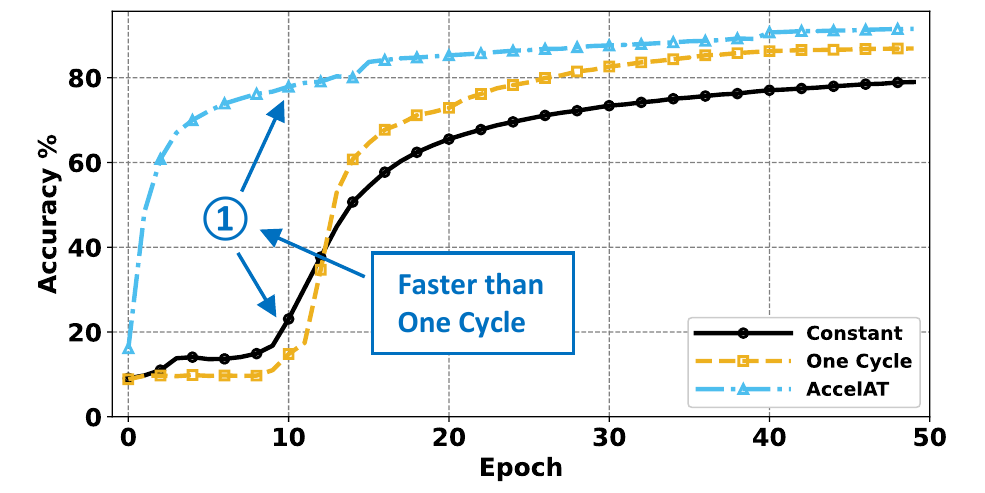}
    \caption{MobileNET model trained on the CIFAR10 dataset, attacked with LinfPGD - Adversarial training-set.}
    \label{fig:figure_mobile_linf_10}
\end{figure}

\begin{figure}[ht]
    \centering
    \includegraphics[width=1\linewidth]{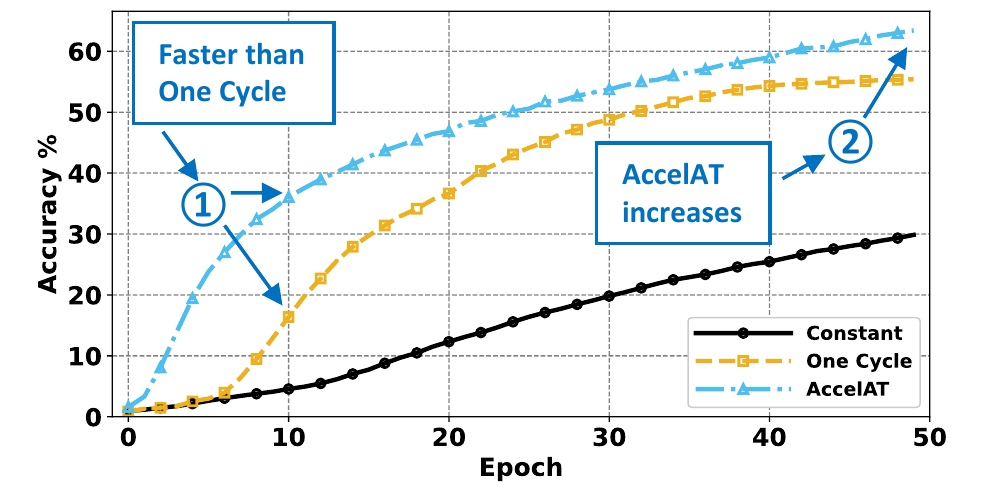}
    \caption{MobileNET model trained on the CIFAR100 dataset, attacked with FGSM - Adversarial training-set.}
    \label{fig:figure_mobile_deep_100}
\end{figure}

From the 36 training runs, we extract nine results that are reported in \cref{fig:figure_res_deep_10}, \cref{fig:figure_mobile_linf_10}, and \cref{fig:figure_mobile_deep_100}. We have shown only the graphs obtained on the adversarial training-set for greater clarity. We then calculated the accuracy with the adversarial test-set, the natural training-set, and the natural test-set. In all cases, the results are comparable to these graphs, with linearly higher or lower values for each type of LR. Note, our \textit{AccelAT} technique is sometimes more efficient than the one cycle policy, especially with MobileNet DNNs.
The best results are obtained on more complex datasets, where many labels reduce the risk of overfitting and where a variable LR is more valuable than a constant one. 
For example, in \cref{fig:figure_mobile_linf_10}, and \cref{fig:figure_mobile_deep_100}, the training curve using the \textit{AccelAT} methodology reaches a high accuracy value in fewer cycles compared to the other techniques (see pointer~1 - \cref{fig:figure_mobile_linf_10}, pointer~1 - \cref{fig:figure_mobile_deep_100}). Also notice that, in \cref{fig:figure_mobile_deep_100}, the accuracy of the curve employing the \textit{AccelAT} technique continues to increase in every cycle, also towards the end of the training (see pointer~2 - \cref{fig:figure_mobile_deep_100}), where instead the accuracy obtained with the one cycle policy stops growing.

\section{Conclusion}
\label{sec:conclusion}

In an era in which robustness is fundamental for DNNs, performing training in a fast yet robust manner is challenging. This research demonstrates that advanced fast training techniques can also be applied to adversarial training, obtaining significant improvements with similar robustness. Moreover, we propose \textit{AccelAT}, a framework for adjusting the LR during training based on the accuracy gradient. Our experimental results show that the \textit{AccelAT} not only outperforms the training with constant LR, which usually reaches a sub-optimal local minimum, but is comparable or better than other fast training techniques. Moreover, it is efficient for complex training with large datasets, where a LR varied at run-time is essential to better fit the DNN model and allows more effective learning. 
Therefore, a new generation of DNN models, which are robust and fast, is starting up, and we demonstrated their feasibility. We believe that, in the near future, these models will be widely employed.

In future works, we plan to modify more hyperparameters by combining them to test even faster methods. Since the batch size is generally fixed and depends on the computing power of the GPU, it affects  less the performance. On the other hand, the momentum can act simultaneously with the LR for better normalization. Hyperparameters are often neglected, but their correct use can lead to significant advantages.

\bibliographystyle{unsrt}
\bibliography{Arxiv2022}

\begin{thebibliography}{10}

\bibitem{DBLP:journals/tccn/Simeone18}
Osvaldo Simeone.
\newblock A very brief introduction to machine learning with applications to
  communication systems.
\newblock {\em {IEEE} Trans. Cogn. Commun. Netw.}, 4(4):648--664, 2018.

\bibitem{DBLP:journals/corr/abs-1708-08296}
Wojciech Samek, Thomas Wiegand, and Klaus{-}Robert M{\"{u}}ller.
\newblock Explainable artificial intelligence: Understanding, visualizing and
  interpreting deep learning models.
\newblock {\em CoRR}, abs/1708.08296, 2017.

\bibitem{ml_raschka_2017}
Sebastian Raschka and Vahid Mirjalili.
\newblock {\em Python Machine Learning}.
\newblock Packt Publishing, 2017.

\bibitem{DBLP:conf/iclr/MadryMSTV18}
Aleksander Madry, Aleksandar Makelov, Ludwig Schmidt, Dimitris Tsipras, and
  Adrian Vladu.
\newblock Towards deep learning models resistant to adversarial attacks.
\newblock In {\em 6th International Conference on Learning Representations,
  {ICLR} 2018, Vancouver, BC, Canada, April 30 - May 3, 2018, Conference Track
  Proceedings}. OpenReview.net, 2018.

\bibitem{Goodfellow-et-al-2016}
Ian Goodfellow, Yoshua Bengio, and Aaron Courville.
\newblock {\em Deep Learning}.
\newblock MIT Press, 2016.

\bibitem{DBLP:conf/nips/ShafahiNG0DSDTG19}
Ali Shafahi, Mahyar Najibi, Amin Ghiasi, Zheng Xu, John~P. Dickerson, Christoph
  Studer, Larry~S. Davis, Gavin Taylor, and Tom Goldstein.
\newblock Adversarial training for free!
\newblock In Hanna~M. Wallach, Hugo Larochelle, Alina Beygelzimer, Florence
  d'Alch{\'{e}}{-}Buc, Emily~B. Fox, and Roman Garnett, editors, {\em Advances
  in Neural Information Processing Systems 32: Annual Conference on Neural
  Information Processing Systems 2019, NeurIPS 2019, December 8-14, 2019,
  Vancouver, BC, Canada}, pages 3353--3364, 2019.

\bibitem{DBLP:conf/nips/ZhangZLZ019}
Dinghuai Zhang, Tianyuan Zhang, Yiping Lu, Zhanxing Zhu, and Bin Dong.
\newblock You only propagate once: Accelerating adversarial training via
  maximal principle.
\newblock In Hanna~M. Wallach, Hugo Larochelle, Alina Beygelzimer, Florence
  d'Alch{\'{e}}{-}Buc, Emily~B. Fox, and Roman Garnett, editors, {\em Advances
  in Neural Information Processing Systems 32: Annual Conference on Neural
  Information Processing Systems 2019, NeurIPS 2019, December 8-14, 2019,
  Vancouver, BC, Canada}, pages 227--238, 2019.

\bibitem{DBLP:conf/icml/ZhangYJXGJ19}
Hongyang Zhang, Yaodong Yu, Jiantao Jiao, Eric~P. Xing, Laurent~El Ghaoui, and
  Michael~I. Jordan.
\newblock Theoretically principled trade-off between robustness and accuracy.
\newblock In Kamalika Chaudhuri and Ruslan Salakhutdinov, editors, {\em
  Proceedings of the 36th International Conference on Machine Learning, {ICML}
  2019, 9-15 June 2019, Long Beach, California, {USA}}, volume~97 of {\em
  Proceedings of Machine Learning Research}, pages 7472--7482. {PMLR}, 2019.

\bibitem{DBLP:journals/corr/abs-1803-09820}
Leslie~N. Smith.
\newblock A disciplined approach to neural network hyper-parameters: Part 1 -
  learning rate, batch size, momentum, and weight decay.
\newblock {\em CoRR}, abs/1803.09820, 2018.

\bibitem{DBLP:conf/iclr/LoshchilovH17}
Ilya Loshchilov and Frank Hutter.
\newblock {SGDR:} stochastic gradient descent with warm restarts.
\newblock In {\em 5th International Conference on Learning Representations,
  {ICLR} 2017, Toulon, France, April 24-26, 2017, Conference Track
  Proceedings}. OpenReview.net, 2017.

\bibitem{DBLP:conf/tencon/MishraS19}
Purnendu Mishra and Kishor Sarawadekar.
\newblock Polynomial learning rate policy with warm restart for deep neural
  network.
\newblock In {\em {TENCON} 2019 - 2019 {IEEE} Region 10 Conference (TENCON),
  Kochi, India, October 17-20, 2019}, pages 2087--2092. {IEEE}, 2019.

\bibitem{DBLP:journals/corr/abs-1810-09619}
Yiwen Guo, Chao Zhang, Changshui Zhang, and Yurong Chen.
\newblock Sparse dnns with improved adversarial robustness.
\newblock {\em CoRR}, abs/1810.09619, 2018.

\bibitem{DBLP:conf/cikm/RuanY021}
Wenjie Ruan, Xinping Yi, and Xiaowei Huang.
\newblock Adversarial robustness of deep learning: Theory, algorithms, and
  applications.
\newblock In Gianluca Demartini, Guido Zuccon, J.~Shane Culpepper, Zi~Huang,
  and Hanghang Tong, editors, {\em {CIKM} '21: The 30th {ACM} International
  Conference on Information and Knowledge Management, Virtual Event,
  Queensland, Australia, November 1 - 5, 2021}, pages 4866--4869. {ACM}, 2021.

\bibitem{DBLP:conf/iclr/KurakinGB17a}
Alexey Kurakin, Ian~J. Goodfellow, and Samy Bengio.
\newblock Adversarial examples in the physical world.
\newblock In {\em 5th International Conference on Learning Representations,
  {ICLR} 2017, Toulon, France, April 24-26, 2017, Workshop Track Proceedings}.
  OpenReview.net, 2017.

\bibitem{DBLP:journals/corr/GoodfellowSS14}
Ian~J. Goodfellow, Jonathon Shlens, and Christian Szegedy.
\newblock Explaining and harnessing adversarial examples.
\newblock In Yoshua Bengio and Yann LeCun, editors, {\em 3rd International
  Conference on Learning Representations, {ICLR} 2015, San Diego, CA, USA, May
  7-9, 2015, Conference Track Proceedings}, 2015.

\bibitem{DBLP:journals/corr/abs-2112-00639}
Nathan Drenkow, Numair Sani, Ilya Shpitser, and Mathias Unberath.
\newblock Robustness in deep learning for computer vision: Mind the gap?
\newblock {\em CoRR}, abs/2112.00639, 2021.

\bibitem{DBLP:journals/csr/PitropakisPGAL19}
Nikolaos Pitropakis, Emmanouil Panaousis, Thanassis Giannetsos, Eleftherios
  Anastasiadis, and George Loukas.
\newblock A taxonomy and survey of attacks against machine learning.
\newblock {\em Comput. Sci. Rev.}, 34, 2019.

\bibitem{DBLP:journals/tec/SuVS19}
Jiawei Su, Danilo~Vasconcellos Vargas, and Kouichi Sakurai.
\newblock One pixel attack for fooling deep neural networks.
\newblock {\em {IEEE} Trans. Evol. Comput.}, 23(5):828--841, 2019.

\bibitem{DBLP:conf/ccs/PapernotMGJCS17}
Nicolas Papernot, Patrick~D. McDaniel, Ian~J. Goodfellow, Somesh Jha, Z.~Berkay
  Celik, and Ananthram Swami.
\newblock Practical black-box attacks against machine learning.
\newblock In Ramesh Karri, Ozgur Sinanoglu, Ahmad{-}Reza Sadeghi, and Xun Yi,
  editors, {\em Proceedings of the 2017 {ACM} on Asia Conference on Computer
  and Communications Security, AsiaCCS 2017, Abu Dhabi, United Arab Emirates,
  April 2-6, 2017}, pages 506--519. {ACM}, 2017.

\bibitem{DBLP:journals/tnn/YuanHZL19}
Xiaoyong Yuan, Pan He, Qile Zhu, and Xiaolin Li.
\newblock Adversarial examples: Attacks and defenses for deep learning.
\newblock {\em {IEEE} Trans. Neural Networks Learn. Syst.}, 30(9):2805--2824,
  2019.

\bibitem{DBLP:conf/nips/SchmidtSTTM18}
Ludwig Schmidt, Shibani Santurkar, Dimitris Tsipras, Kunal Talwar, and
  Aleksander Madry.
\newblock Adversarially robust generalization requires more data.
\newblock In Samy Bengio, Hanna~M. Wallach, Hugo Larochelle, Kristen Grauman,
  Nicol{\`{o}} Cesa{-}Bianchi, and Roman Garnett, editors, {\em Advances in
  Neural Information Processing Systems 31: Annual Conference on Neural
  Information Processing Systems 2018, NeurIPS 2018, December 3-8, 2018,
  Montr{\'{e}}al, Canada}, pages 5019--5031, 2018.

\bibitem{DBLP:journals/nn/EsfandiariBEVES21}
Yasaman Esfandiari, Aditya Balu, Keivan Ebrahimi, Umesh Vaidya, Nicola Elia,
  and Soumik Sarkar.
\newblock A fast saddle-point dynamical system approach to robust deep
  learning.
\newblock {\em Neural Networks}, 139:33--44, 2021.

\bibitem{DBLP:conf/iclr/ShafahiSZGSJG20}
Ali Shafahi, Parsa Saadatpanah, Chen Zhu, Amin Ghiasi, Christoph Studer,
  David~W. Jacobs, and Tom Goldstein.
\newblock Adversarially robust transfer learning.
\newblock In {\em 8th International Conference on Learning Representations,
  {ICLR} 2020, Addis Ababa, Ethiopia, April 26-30, 2020}. OpenReview.net, 2020.

\bibitem{DBLP:conf/nips/GuiWYYW019}
Shupeng Gui, Haotao Wang, Haichuan Yang, Chen Yu, Zhangyang Wang, and Ji~Liu.
\newblock Model compression with adversarial robustness: {A} unified
  optimization framework.
\newblock In Hanna~M. Wallach, Hugo Larochelle, Alina Beygelzimer, Florence
  d'Alch{\'{e}}{-}Buc, Emily~B. Fox, and Roman Garnett, editors, {\em Advances
  in Neural Information Processing Systems 32: Annual Conference on Neural
  Information Processing Systems 2019, NeurIPS 2019, December 8-14, 2019,
  Vancouver, BC, Canada}, pages 1283--1294, 2019.

\bibitem{DBLP:conf/iclr/MetzenGFB17}
Jan~Hendrik Metzen, Tim Genewein, Volker Fischer, and Bastian Bischoff.
\newblock On detecting adversarial perturbations.
\newblock In {\em 5th International Conference on Learning Representations,
  {ICLR} 2017, Toulon, France, April 24-26, 2017, Conference Track
  Proceedings}. OpenReview.net, 2017.

\bibitem{DBLP:conf/cvpr/Moosavi-Dezfooli17}
Seyed{-}Mohsen Moosavi{-}Dezfooli, Alhussein Fawzi, Omar Fawzi, and Pascal
  Frossard.
\newblock Universal adversarial perturbations.
\newblock In {\em 2017 {IEEE} Conference on Computer Vision and Pattern
  Recognition, {CVPR} 2017, Honolulu, HI, USA, July 21-26, 2017}, pages 86--94.
  {IEEE} Computer Society, 2017.

\bibitem{DBLP:conf/iclr/PangYDSZ21}
Tianyu Pang, Xiao Yang, Yinpeng Dong, Hang Su, and Jun Zhu.
\newblock Bag of tricks for adversarial training.
\newblock In {\em 9th International Conference on Learning Representations,
  {ICLR} 2021, Virtual Event, Austria, May 3-7, 2021}. OpenReview.net, 2021.

\bibitem{DBLP:journals/corr/abs-2010-03593}
Sven Gowal, Chongli Qin, Jonathan Uesato, Timothy~A. Mann, and Pushmeet Kohli.
\newblock Uncovering the limits of adversarial training against norm-bounded
  adversarial examples.
\newblock {\em CoRR}, abs/2010.03593, 2020.

\bibitem{DBLP:journals/corr/GoodfellowPM16}
Ian~J. Goodfellow, Nicolas Papernot, and Patrick~D. McDaniel.
\newblock cleverhans v0.1: an adversarial machine learning library.
\newblock {\em CoRR}, abs/1610.00768, 2016.

\bibitem{DBLP:journals/corr/abs-2001-05574}
Dou Goodman, Xin Hao, Yang Wang, Yuesheng Wu, Junfeng Xiong, and Huan Zhang.
\newblock Advbox: a toolbox to generate adversarial examples that fool neural
  networks.
\newblock {\em CoRR}, abs/2001.05574, 2020.

\bibitem{DBLP:journals/corr/RauberBB17}
Jonas Rauber, Wieland Brendel, and Matthias Bethge.
\newblock Foolbox v0.8.0: {A} python toolbox to benchmark the robustness of
  machine learning models.
\newblock {\em CoRR}, abs/1707.04131, 2017.

\bibitem{DBLP:journals/corr/AbadiABBCCCDDDG16}
Mart{\'{\i}}n Abadi, Ashish Agarwal, Paul Barham, Eugene Brevdo, Zhifeng Chen,
  Craig Citro, Gregory~S. Corrado, Andy Davis, Jeffrey Dean, Matthieu Devin,
  Sanjay Ghemawat, Ian~J. Goodfellow, Andrew Harp, Geoffrey Irving, Michael
  Isard, Yangqing Jia, Rafal J{\'{o}}zefowicz, Lukasz Kaiser, Manjunath Kudlur,
  Josh Levenberg, Dan Man{\'{e}}, Rajat Monga, Sherry Moore, Derek~Gordon
  Murray, Chris Olah, Mike Schuster, Jonathon Shlens, Benoit Steiner, Ilya
  Sutskever, Kunal Talwar, Paul~A. Tucker, Vincent Vanhoucke, Vijay Vasudevan,
  Fernanda~B. Vi{\'{e}}gas, Oriol Vinyals, Pete Warden, Martin Wattenberg,
  Martin Wicke, Yuan Yu, and Xiaoqiang Zheng.
\newblock Tensorflow: Large-scale machine learning on heterogeneous distributed
  systems.
\newblock {\em CoRR}, abs/1603.04467, 2016.

\bibitem{DBLP:conf/nips/PaszkeGMLBCKLGA19}
Adam Paszke, Sam Gross, Francisco Massa, Adam Lerer, James Bradbury, Gregory
  Chanan, Trevor Killeen, Zeming Lin, Natalia Gimelshein, Luca Antiga, Alban
  Desmaison, Andreas K{\"{o}}pf, Edward~Z. Yang, Zachary DeVito, Martin Raison,
  Alykhan Tejani, Sasank Chilamkurthy, Benoit Steiner, Lu~Fang, Junjie Bai, and
  Soumith Chintala.
\newblock Pytorch: An imperative style, high-performance deep learning library.
\newblock In Hanna~M. Wallach, Hugo Larochelle, Alina Beygelzimer, Florence
  d'Alch{\'{e}}{-}Buc, Emily~B. Fox, and Roman Garnett, editors, {\em Advances
  in Neural Information Processing Systems 32: Annual Conference on Neural
  Information Processing Systems 2019, NeurIPS 2019, December 8-14, 2019,
  Vancouver, BC, Canada}, pages 8024--8035, 2019.

\bibitem{DBLP:journals/corr/KingmaB14}
Diederik~P. Kingma and Jimmy Ba.
\newblock Adam: {A} method for stochastic optimization.
\newblock In Yoshua Bengio and Yann LeCun, editors, {\em 3rd International
  Conference on Learning Representations, {ICLR} 2015, San Diego, CA, USA, May
  7-9, 2015, Conference Track Proceedings}, 2015.

\bibitem{song2019adaboundary}
Hwanjun Song, Sundong Kim, Minseok Kim, and Jae-Gil Lee.
\newblock Ada-boundary: Accelerating the {DNN} training via adaptive boundary
  batch selection, 2019.

\bibitem{DBLP:journals/corr/abs-1708-07120}
Leslie~N. Smith and Nicholay Topin.
\newblock Super-convergence: Very fast training of residual networks using
  large learning rates.
\newblock {\em CoRR}, abs/1708.07120, 2017.

\bibitem{DBLP:conf/wacv/Smith17}
Leslie~N. Smith.
\newblock Cyclical learning rates for training neural networks.
\newblock In {\em 2017 {IEEE} Winter Conference on Applications of Computer
  Vision, {WACV} 2017, Santa Rosa, CA, USA, March 24-31, 2017}, pages 464--472.
  {IEEE} Computer Society, 2017.

\bibitem{DBLP:journals/corr/abs-1712-02029}
Aditya Devarakonda, Maxim Naumov, and Michael Garland.
\newblock Adabatch: Adaptive batch sizes for training deep neural networks.
\newblock {\em CoRR}, abs/1712.02029, 2017.

\bibitem{DBLP:conf/iclr/WongRK20}
Eric Wong, Leslie Rice, and J.~Zico Kolter.
\newblock Fast is better than free: Revisiting adversarial training.
\newblock In {\em 8th International Conference on Learning Representations,
  {ICLR} 2020, Addis Ababa, Ethiopia, April 26-30, 2020}. OpenReview.net, 2020.

\bibitem{DBLP:conf/nips/AndriushchenkoF20}
Maksym Andriushchenko and Nicolas Flammarion.
\newblock Understanding and improving fast adversarial training.
\newblock In Hugo Larochelle, Marc'Aurelio Ranzato, Raia Hadsell,
  Maria{-}Florina Balcan, and Hsuan{-}Tien Lin, editors, {\em Advances in
  Neural Information Processing Systems 33: Annual Conference on Neural
  Information Processing Systems 2020, NeurIPS 2020, December 6-12, 2020,
  virtual}, 2020.

\bibitem{DBLP:conf/cvpr/HeZRS16}
Kaiming He, Xiangyu Zhang, Shaoqing Ren, and Jian Sun.
\newblock Deep residual learning for image recognition.
\newblock In {\em 2016 {IEEE} Conference on Computer Vision and Pattern
  Recognition, {CVPR} 2016, Las Vegas, NV, USA, June 27-30, 2016}, pages
  770--778. {IEEE} Computer Society, 2016.

\bibitem{cifar10}
Alex Krizhevsky.
\newblock Learning multiple layers of features from tiny images.
\newblock Technical report, Department of Computer Science, University of
  Toronto, 2009.

\bibitem{Krizhevsky09learningmultiple}
Alex Krizhevsky.
\newblock Learning multiple layers of features from tiny images.
\newblock Technical report, University of Toronto, 2009.

\bibitem{cifar100}
Alex Krizhevsky.
\newblock Learning multiple layers of features from tiny images.
\newblock Technical report, Department of Computer Science, University of
  Toronto, 2009.

\bibitem{DBLP:conf/icml/LiuY09}
Jun Liu and Jieping Ye.
\newblock Efficient euclidean projections in linear time.
\newblock In Andrea~Pohoreckyj Danyluk, L{\'{e}}on Bottou, and Michael~L.
  Littman, editors, {\em Proceedings of the 26th Annual International
  Conference on Machine Learning, {ICML} 2009, Montreal, Quebec, Canada, June
  14-18, 2009}, volume 382 of {\em {ACM} International Conference Proceeding
  Series}, pages 657--664. {ACM}, 2009.

\bibitem{DBLP:journals/corr/HowardZCKWWAA17}
Andrew~G. Howard, Menglong Zhu, Bo~Chen, Dmitry Kalenichenko, Weijun Wang,
  Tobias Weyand, Marco Andreetto, and Hartwig Adam.
\newblock Mobilenets: Efficient convolutional neural networks for mobile vision
  applications.
\newblock {\em CoRR}, abs/1704.04861, 2017.

\bibitem{DBLP:journals/information/RaschkaPN20}
Sebastian Raschka, Joshua Patterson, and Corey Nolet.
\newblock Machine learning in python: Main developments and technology trends
  in data science, machine learning, and artificial intelligence.
\newblock {\em Inf.}, 11(4):193, 2020.

\bibitem{DBLP:conf/icml/RiceWK20}
Leslie Rice, Eric Wong, and J.~Zico Kolter.
\newblock Overfitting in adversarially robust deep learning.
\newblock In {\em Proceedings of the 37th International Conference on Machine
  Learning, {ICML} 2020, 13-18 July 2020, Virtual Event}, volume 119 of {\em
  Proceedings of Machine Learning Research}, pages 8093--8104. {PMLR}, 2020.

\bibitem{DBLP:conf/icml/ZhangXH0CSK20}
Jingfeng Zhang, Xilie Xu, Bo~Han, Gang Niu, Lizhen Cui, Masashi Sugiyama, and
  Mohan~S. Kankanhalli.
\newblock Attacks which do not kill training make adversarial learning
  stronger.
\newblock In {\em Proceedings of the 37th International Conference on Machine
  Learning, {ICML} 2020, 13-18 July 2020, Virtual Event}, volume 119 of {\em
  Proceedings of Machine Learning Research}, pages 11278--11287. {PMLR}, 2020.

\bibitem{DBLP:journals/corr/abs-1902-06705}
Nicholas Carlini, Anish Athalye, Nicolas Papernot, Wieland Brendel, Jonas
  Rauber, Dimitris Tsipras, Ian~J. Goodfellow, Aleksander Madry, and Alexey
  Kurakin.
\newblock On evaluating adversarial robustness.
\newblock {\em CoRR}, abs/1902.06705, 2019.

\end{thebibliography}

\clearpage

\begin{IEEEbiography}[{\includegraphics[width=1in,height=1.25in,clip,keepaspectratio]{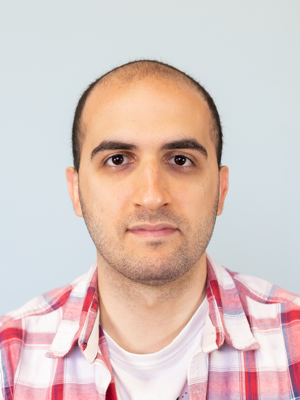}}]{Farzad Nikfam} (Graduate Student Member, IEEE) received the Bachelor’s degree in Mechanical Engineering, in March 2018, and the Master’s degree in Mechatronic Engineering, in March 2020 from Politecnico di Torino, Turin, Italy. Now he is pursuing a PhD in Electronics and Communication Engineering under the supervision of Prof. Maurizio Martina at Politecnico di Torino. His research activity is focused on software machine learning with emphasis on security and privacy problems. He is an IEEE student member since 2021.
\end{IEEEbiography}

\begin{IEEEbiography}[{\includegraphics[width=1in,height=1.25in,clip,keepaspectratio]{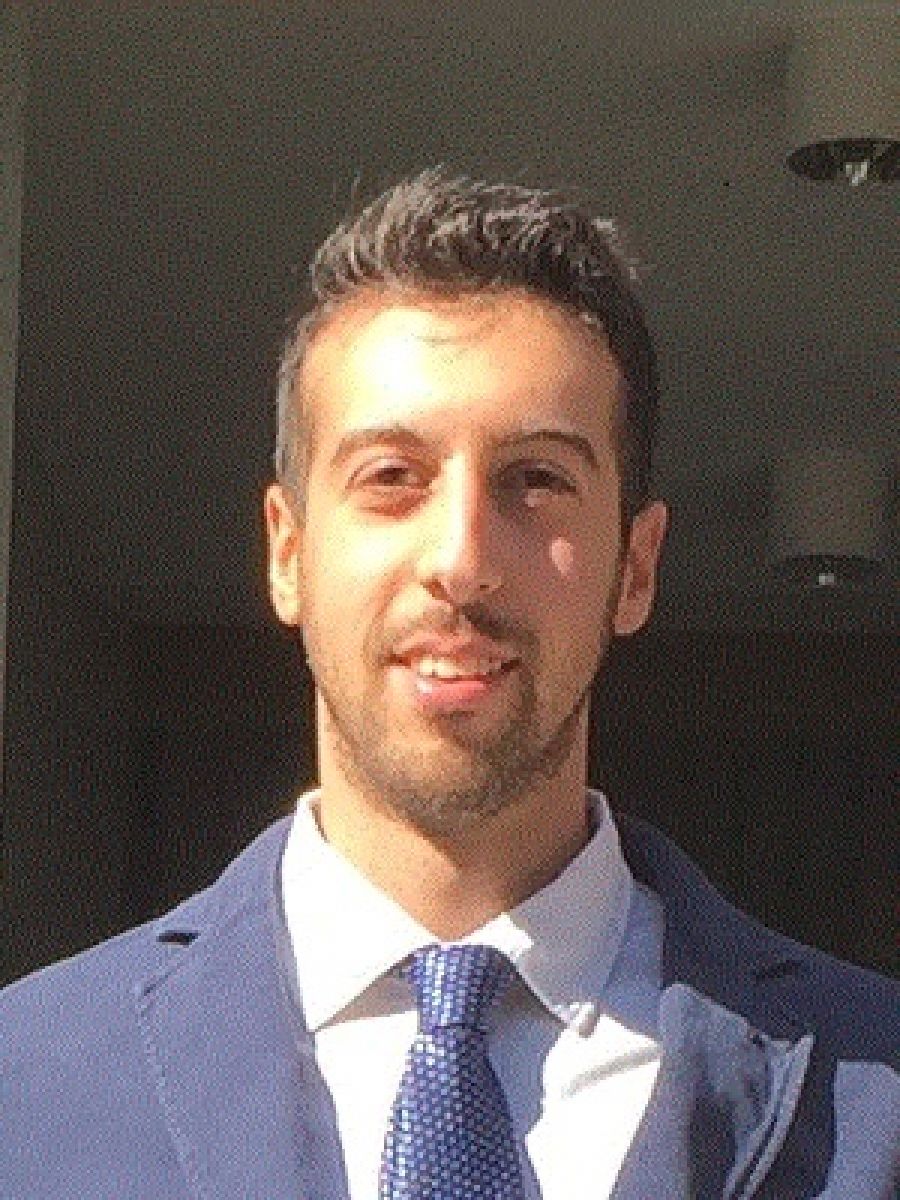}}]{Alberto Marchisio} (Graduate Student Member, IEEE) received his B.Sc. and M.Sc. degrees in Electronic Engineering from Politecnico di Torino, Turin, Italy, in October 2015 and April 2018, respectively. Currently, he is Ph.D. Student at Computer Architecture and Robust Energy-Efficient Technologies (CARE-Tech.) lab, Institute of Computer Engineering, Technische Universit{\"a}t Wien (TU Wien), Vienna, Austria, under the supervision of Prof. Dr. Muhammad Shafique. His main research interests include hardware and software optimizations for machine learning, brain-inspired computing, VLSI architecture design, emerging computing technologies, robust design, and approximate computing for energy efficiency. He (co-)authored 20+ papers in prestigious international conferences and journals. He received the honorable mention at the Italian National Finals of Maths Olympic Games in 2012, and the Richard Newton Young Fellow Award in 2019.
\end{IEEEbiography}

\begin{IEEEbiography}[{\includegraphics[width=1in,height=1.25in,clip,keepaspectratio]{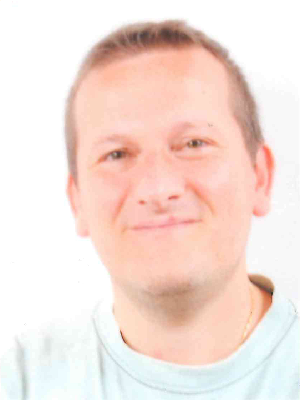}}]{Maurizio Martina} (Senior Member, IEEE) received the M.S. and Ph.D. in electrical
engineering from Politecnico di Torino, Italy, in 2000 and 2004, respectively. He is currently Full Professor with the VLSI-Lab group, Politecnico di Torino. His research interests include computer architecture and VLSI design of architectures for digital signal processing, video coding, communications, networking, artificial intelligence, machine learning and event-based processing. He edited one book and published 3 book chapters on VLSI architectures and digital circuits for video coding, wireless communications and error correcting codes. He has more than 100 scientific publications and is co-author of 2 patents. He is now an Associate Editor of IEEE Transactions on Circuits and Systems - I. He had been part of the organizing and technical committee of several international conferences, including BioCAS 2017, ICECS 2019, AICAS 2020. Currently, he is the counselor of the IEEE Student Branch at Politecnico di Torino and a professional member of IEEE HKN.
\end{IEEEbiography}

\begin{IEEEbiography}[{\includegraphics[width=1in,height=1.25in,clip,keepaspectratio]{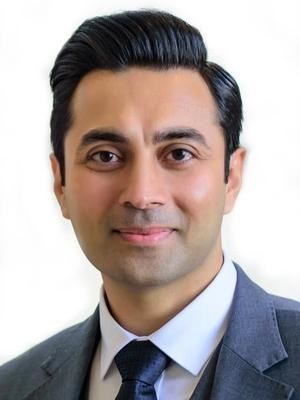}}]{Muhammad Shafique} (Senior Member, IEEE) received the Ph.D. degree in computer science from the Karlsruhe Institute of Technology (KIT), Germany, in 2011. Afterwards, he established and led a highly recognized research group at KIT for several years as well as conducted impactful collaborative R\&D activities across the globe. In Oct.2016, he joined the Institute of Computer Engineering at the Faculty of Informatics, Technische Universität Wien (TU Wien), Vienna, Austria as a Full Professor of Computer Architecture and Robust, Energy-Efficient Technologies. Since Sep.2020, Dr. Shafique is with the New York University (NYU), where he is currently a Full Professor and the director of the eBrain Lab at the NYU-Abu Dhabi in UAE, and a Global Network Professor at the Tandon School of Engineering, NYU-New York City in USA. He is also a Co-PI/Investigator in multiple NYUAD Centers, including Center of Artificial Intelligence and Robotics (CAIR), Center of Cyber Security (CCS), Center for InTeractIng urban nEtworkS (CITIES), and Center for Quantum and Topological Systems (CQTS). His research interests are in AI \& machine learning hardware and system-level design, brain-inspired computing, autonomous systems, wearable healthcare, energy-efficient systems, robust computing, hardware security, emerging technologies, FPGAs, MPSoCs, and embedded systems. His research has a special focus on cross-layer analysis, modeling, design, and optimization of computing and memory systems. The researched technologies and tools are deployed in application use cases from Internet-of-Things (IoT), smart Cyber-Physical Systems (CPS), and ICT for Development (ICT4D) domains. Dr. Shafique has given several Keynotes, Invited Talks, and Tutorials, as well as organized many special sessions at premier venues. He has served as the PC Chair, General Chair, Track Chair, and PC member for several prestigious IEEE/ACM conferences. Dr. Shafique holds one U.S. patent has (co-)authored 6 Books, 15+ Book Chapters, 350+ papers in premier journals and conferences, and 50+ archive articles. He received the 2015 ACM/SIGDA Outstanding New Faculty Award, the AI 2000 Chip Technology Most Influential Scholar Award in 2020 and 2022, the ASPIRE AARE Research Excellence Award in 2021, six gold medals, and several best paper awards and nominations at prestigious conferences. He is a senior member of the IEEE and IEEE Signal Processing Society (SPS), and a member of the ACM, SIGARCH, SIGDA, SIGBED, and HIPEAC.
\end{IEEEbiography}

\EOD

\end{document}